\RequirePackage{fix-cm}
\documentclass[final,5p,times,twocolumn]{elsarticle}

\usepackage{url}
\usepackage{microtype}

\usepackage{amsmath} 
\usepackage{amssymb}  
\usepackage{array}
\usepackage[linesnumbered]{algorithm2e}
\usepackage{float} 
\usepackage{siunitx}

\usepackage{graphicx}
\usepackage{caption}
\usepackage{subcaption}
\usepackage[dvipsnames]{xcolor}
\usepackage{xcolor,colortbl}

\usepackage{enumitem,booktabs}
\usepackage{multirow}
\usepackage[referable]{threeparttablex}
\renewlist{tablenotes}{enumerate}{1}
\usepackage[colorlinks=true,allcolors=blue]{hyperref}

\usepackage[official]{eurosym}

\makeatletter
\setlist[tablenotes]{label=\tnote{\alph*},ref=\alph*,itemsep=\z@,topsep=\z@skip,partopsep=\z@skip,parsep=\z@,itemindent=\z@,labelindent=\tabcolsep,labelsep=.2em,leftmargin=*,align=left,before={\footnotesize}}
\makeatother

\newcommand\mystrut[1][2]{%
    \setlength\mylena{#1\ht\@arstrutbox}%
    \setlength\mylenb{#1\dp\@arstrutbox}%
    \rule[\mylenb]{0pt}{\mylena}}
\makeatother

\newcommand{\argmax}[1]{\underset{#1}{\operatorname{arg}\,\operatorname{max}}\;}
\newcommand{\argmin}[1]{\underset{#1}{\operatorname{arg}\,\operatorname{min}}\;}

\usepackage{color}

\begin{document}
\journal{Robotics and Autonomous Systems}

\begin{frontmatter}

\title{Active Robot-Assisted Feeding with a General-Purpose Mobile Manipulator: Design, Evaluation, and Lessons Learned}
\author[hrl]{Daehyung Park\corref{cor1}}
\ead{daehyung@csail.mit.edu}

\author[hrl]{Yuuna Hoshi}
\author[gt]{Harshal P. Mahajan}
\author[hrl]{Ho Keun Kim}
\author[hrl]{Zackory Erickson}
\author[uiuc]{Wendy A. Rogers}
\author[hrl]{Charles C. Kemp}

\address[hrl]{Healthcare Robotics Lab, Georgia Institute of Technology, Atlanta, GA, USA}
\address[gt]{Georgia Institute of Technology, Atlanta, GA, USA}
\address[uiuc]{University of Illinois Urbana-Champaign, Champaign, IL, USA}

\cortext[cor1]{Corresponding author}


\begin{abstract}
Eating is an essential activity of daily living (ADL) for staying healthy and living at home independently. Although numerous assistive devices have been introduced, many people with disabilities are still restricted from independent eating due to the devices' physical or perceptual limitations. In this work, we present a new meal-assistance system and evaluations of this system with people with motor impairments. We also discuss learned lessons and design insights based on the evaluations. The meal-assistance system uses a general-purpose mobile manipulator, a Willow Garage PR2, which has the potential to serve as a versatile form of assistive technology. Our \textit{active feeding} framework enables the robot to autonomously deliver food to the user's mouth, reducing the need for head movement by the user. The user interface, visually-guided behaviors, and safety tools allow people with severe motor impairments to successfully use the system. We evaluated our system with a total of 10 able-bodied participants and 9 participants with motor impairments. Both groups of participants successfully ate various foods using the system and reported high rates of success for the system's autonomous behaviors. In general, participants who operated the system reported that it was comfortable, safe, and easy-to-use. 
\end{abstract}

\begin{keyword}
Assistive Robots \sep Manipulation \sep Assistive Feeding \sep Meal Assistance
\end{keyword}
\end{frontmatter}

\section{Introduction} \label{intro}
Activities of daily living (ADLs), such as eating, toileting, and dressing, are important for quality of life \cite{wiener1990measuring}. Yet for many people with disabilities, including people with upper limb impairments, such tasks prove challenging without assistance from a human caregiver. However, shortages of healthcare workers and rising healthcare costs create a pressing need for innovations that make assistance more affordable and effective.

Technology interventions can be a solution by bridging the gap between physical capability and necessary functional ability \cite{mitzner2018closing}. Numerous specialized assistive devices, including robots, have been developed to help people with disabilities perform ADLs on their own \cite{rashidi2013survey}. Each device typically provides a narrow form of assistance suitable for people with particular impairments. Alternatively, researchers have applied general-purpose mobile manipulators to a variety of applications, such as rescue, assistance, and residential service \cite{schwarz2017nimbro, deegan2008mobile}. The robots often have a mobile base and human-like arms (e.g., PR2 robot from Willow Garage \cite{pr2} and Jaco arm with a mobile base from Fattal et al. \cite{fattal2018sam}), and help people to overcome their physical or perceptual limitations via teleoperation \cite{grice2018home}. Although mobile manipulators have the potential to provide a wide variety of assistive services \cite{chen2013robots}, their complexity creates challenges, including the risk of low usability.

\begin{figure}[t]
	\centering
    \includegraphics[trim={0cm, 0cm, 0cm, 0cm}, clip, width=85mm]{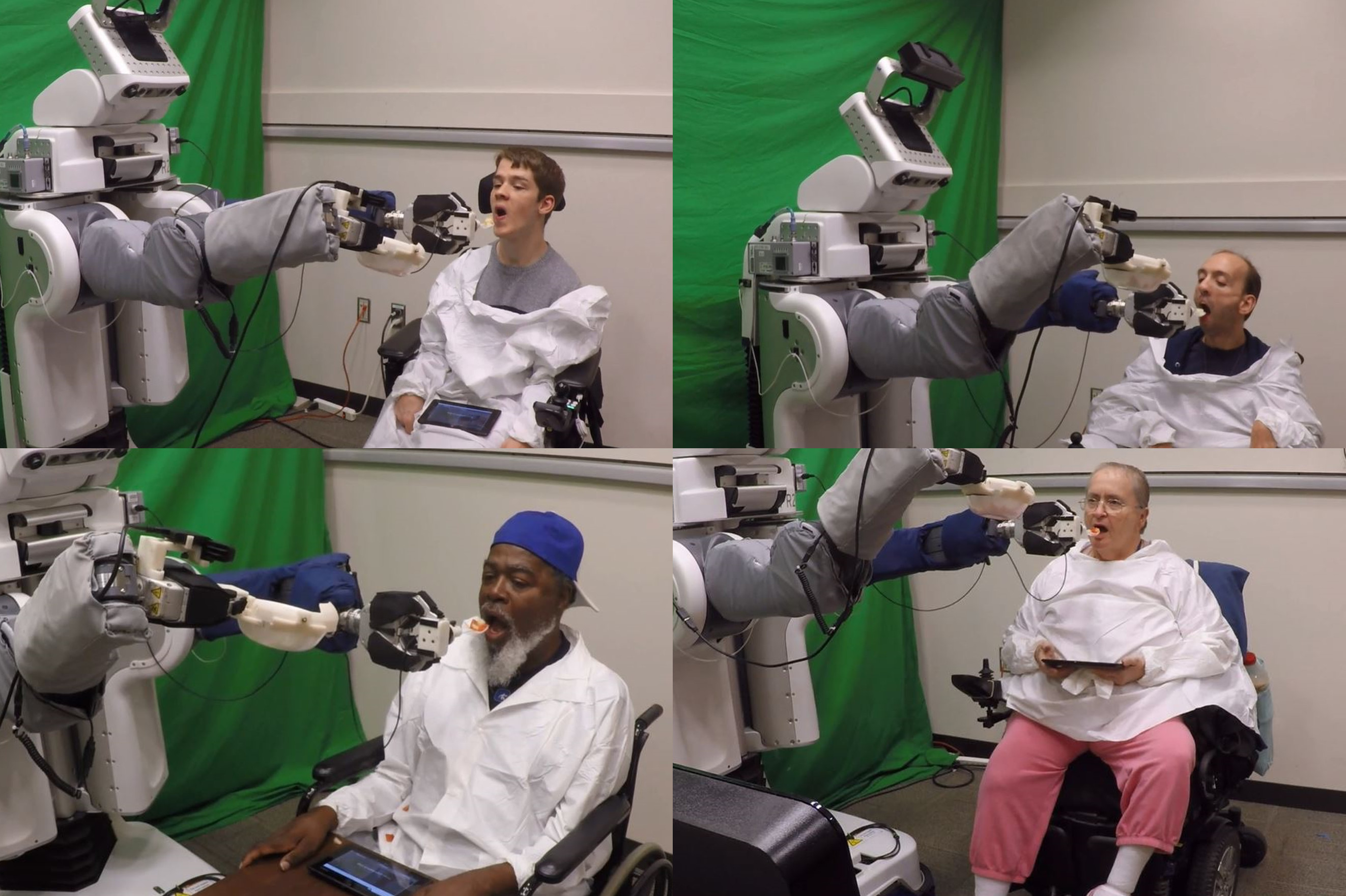}    
	\caption{People with motor impairments eat food using the robot-assisted feeding system at the Healthcare Robotics Lab at Georgia Tech.}    
	\label{fig:feeding_shot}
\end{figure}

A representative assistive task is meal assistance, which is an essential ADL for staying healthy. People with upper-body and limb impairments often have difficulty feeding themselves. Although a number of specialized meal-assistance robots are commercially available (e.g., My Spoon \cite{myspoon}, Bestic arm \cite{bestic}, and Mealtime partner \cite{mealtime}), these robots provide limited meal assistance. Notably, we refer to the type of assistance these robots provide as \textit{passive feeding} assistance, where the robot delivers food to a predefined location outside the users' mouth and users take the food by using their upper body and limb movement. This is due in part to the robots' (desk-mountable) fixed bases, low degree-of-freedom (DoF) arms, and limited sensing capabilities. Instead, we use a general-purpose mobile manipulator to provide \textit{active feeding} assistance that autonomously delivers food inside a user's mouth, taking advantage of the robot's greater physical and sensing capabilities.

In this paper, we introduce a meal-assistance system that enables a general-purpose mobile manipulator, a PR2 robot, to provide safe, easy-to-use assistance with feeding (see Fig.~\ref{fig:feeding_shot}). The system provides \textit{active feeding} assistance in which the PR2 uses visually-guided movements to autonomously scoop/stab food and deliver the food inside a user's mouth. The system also provides a graphical user interface (GUI) for people with various motor impairments to easily command three independent subtasks: scooping/stabbing (food acquisition), spoon-wiping (removing excess food), and delivery (feeding). Note that we group the scooping and stabbing subtasks in terms of their similar functionality (i.e., food acquisition), but the subtasks use different tools, motions, and foods. Further, our system provides state-of-the-art safety functions that proactively monitor and prevent potential anomalous executions.

Our primary contribution is that instead of a specialized meal-assistance device, our system uses a general-purpose mobile manipulator to provide \textit{active feeding} assistance that addresses considerations found in the literature: convenience, comfort, speed, and safety as well as food acquisition and delivery functions \cite{tejima1996rehabilitation, song2012novel}. For convenience, our system allows 5 utensils and 2 types of bowls to adapt the functions to the user-selected food.
The system has software and hardware interfaces to allow caregivers to replace the utensil and bowl depending on the type of food. The system also allows users to access its interface from a web browser, enabling the use of a variety of devices and increasing accessibility \cite{grice2018home}. To improve safety, we designed the system to use compliant arm motions with a low-gain controller as well as a multimodal execution monitor to detect and react to anomalous events during assistance \cite{park2017class, park2018multimodal}.

Another contribution is the evaluation of the system with two groups, able-bodied participants and participants with motor impairments, for comparing the design factors. As a step towards use by people with motor impairments, we first evaluated our system with 9 able-bodied participants. We then evaluated the system with 8 people for whom unassisted feeding was difficult due to physical disabilities. We compare the two groups of evaluation results and show the system is safe, convenient, and easy-to-use. 
In addition to these two laboratory studies, the first author performed a long-term self evaluation while developing the system, and we evaluated the system with Henry Evans\footnote{Henry Evans became quadriplegic and mute after a stroke in 2003. As our main collaborator, he has participated in several of our assistive robotics studies since 2010.}, a person with quadriplegia, who operated the system to feed himself at his home. We also discuss learned lessons and design insights toward potential meal-assistance systems for people with motor impairments.

The new and previously unpublished content in the current paper includes the following:
\begin{itemize}
    \item We present a detailed description of an improved meal-assistance system with visually-guided behaviors for autonomous food acquisition and delivery, and an updated graphical user interface (GUI). 
    \item We present a wholly new evaluation of the meal-assistance system with 8 people with disabilities who have difficulty feeding themselves. 
    \item We present new results and analyses based on the new study with 8 people with disabilities, a new long-term self evaluation, and a previous study with 9 able-bodied participants \cite{park2017class}. 
    \item We share learned lessons and design insights for assistive robots.
\end{itemize}

The rest of this paper is organized as follows: Section \ref{sec:2} shows related work including the examples of assistive robots, particularly assistive feeding devices. Section \ref{sec:3} presents the outline of our meal-assistance system. Section \ref{sec:4} describes the individual components of the system. Then, Sections \ref{sec:5} and \ref{sec:6} show our experimental setup and results, respectively. Finally, we present design insights and conclusions in Sections \ref{sec:7} and \ref{sec:8}, respectively.

\section{Related Work} \label{sec:2}
Assistive robots are a type of devices that can provide physical, mental, or social assistance to people with disabilities or seniors \cite{feil2005defining, maciejasz2014survey}. 
In this section, we review assistive robots, particularly manipulators, for ADLs. We then go over meal-assistance devices including feeding robots.

\newcolumntype{K}{>{\centering\arraybackslash}m{0.1cm}<{}}
\newcolumntype{M}{>{\arraybackslash}m{1.9cm}<{}}
\newcolumntype{N}{>{\centering\arraybackslash}m{2.0cm}<{}}
\newcolumntype{O}{>{\centering\arraybackslash}m{1.8cm}<{}}
\definecolor{Gray}{gray}{0.5}

\begin{table*}
\centering
\caption{A survey of recent robot-assisted feeding systems.}
\label{tab:feeding_robots}       
\begin{threeparttable}
    \begin{tabular}{c l M c N O c c}
	\toprule
	& \multirow{2}{*}{Platform} & \multirow{2}{*}{Interface\tnotex{tnote:robots-r1}} & \multirow{2}{*}{Tool\tnotex{tnote:robots-r2}} & \multicolumn{2}{c}{Teaching/Movement Type  } & \multirow{2}{*}{Base} & \multirow{2}{*}{Safety Tool}  \\
    \cmidrule(lr){5-6}
	& & & & scooping & delivery & & \\
	\midrule
    & My Spoon \cite{myspoon} & Joystick & sf & Predefined & - & Fixed & - \\
	& Bestic arm \cite{bestic} & Button & s & Predefined & - & Fixed & - \\
	& Meal Buddy \cite{meal_buddy} & Joystick & s & Predefined & - & Fixed & - \\
	& Mealtime \cite{mealtime} & Button & s & Predefined & - & Fixed & A shatterproof spoon \\
	\multirow{-5}{*}{ \rotatebox[origin=c]{90}{Commercial}} & Obi \cite{obi} & Button & s & Predefined & Kinesthetic & Fixed & Collision detection \\
	\arrayrulecolor{Gray}\hline
    \arrayrulecolor{Black}
	& Yamazaki and Masuda \cite{yamazaki2012various} & GUI(H) & sfc & User-selected\tnotex{tnote:robots-r3} & Predefined & Fixed & Force detection \\
	& Song and Kim \cite{song2012novel} & Joystick \mbox{\& Button} & sg & Predefined & Predefined & Fixed & - \\    
	& Schroer et al. \cite{schroer2015autonomous}\tnotex{tnote:robots-r4} & BCI\tnotex{tnote:robots-r5} & N/A & N/A & Vision & Movable & - \\
	& Kobayashi et al. \cite{kobayashi2016meal} & Touch sensor & sc & Vision & Predefined & Fixed & Spring joint \\
	& Perera et al. \cite{perera2016ssvep,perera2017eeg} & BCI & s & Predefined & Predefined & Fixed & - \\
    & Admoni and Srinivasa \cite{admoni2017gaze} & Joystick \& Gaze & f & User-selected & Predefined & Fixed & - \\
    & Candeias et al. \cite{candeias2018feeding} & N/A & s & Predefined &  Vision & Fixed & - \\ 
	\multirow{-10}{*}{\rotatebox[origin=c]{90}{Academic}} & Our Work & GUI(H) & sf & Vision & Vision & Movable & Execution monitor \\
	\bottomrule
	\end{tabular}
    \begin{tablenotes}
      \item\label{tnote:robots-r1}H: A head tracker is used as a pointing device, E: An eye tracker is used as pointing device.
      \item\label{tnote:robots-r2}s: spoon, f: fork, c: chopstick, g: gripper.
      \item\label{tnote:robots-r3}A user manually selects a target location on a screen.
	  \item\label{tnote:robots-r4}Drinking task only.     
      \item\label{tnote:robots-r5}Brain-computer interface (BCI). 
    \end{tablenotes}
\end{threeparttable}    
\end{table*}

\subsection{Our Prior Work on Robot-Assisted Feeding}

This paper represents the culmination of work that we began in 2014 to develop an intelligent meal-assistance system. A workshop publication described an early, less-capable version of the meal-assistance system that required fiducial markers placed on the person's head and the bowl \cite{park2016towards}. Otherwise, our publications involving meal-assistance have focused on execution monitoring \cite{park2016multimodal, park2017class,park2018detection, park2018multimodal}. The newer meal-assistance system that we present now was used in a conference paper \cite{park2017class} to evaluate an execution monitoring system, but the paper provided no details about the meal-assistance system. Prior to our new study with 8 people with disabilities that we present here for the first time, we had only evaluated the system with a single person with disabilities (i.e., Henry Evans) who was involved in the system's development. The 8 participants in our new study had no prior involvement with the system. 

\subsection{Assistive Manipulators}
Researchers have introduced a wide variety of assistive manipulators---such as 7-DoF arms mounted on a wheelchair or desk---to provide general assistance near the human \cite{hammel1989clinical, van1999provar, hillman1999wheelchair, dietsch2006portable, wakita2012user}. We categorize the types of manipulators in terms of mobility: fixed- and mobile-base manipulators.

\paragraph{\textbf{Fixed-base Robots}}
Fixed-base assistive robots are often placed near a user or a targeted workspace. Researchers mounted early assistive robots to desktops for assistance with feeding, cosmetics, and hygiene. The professional vocational assistive robot (ProVAR) is a representative desktop manipulator placed in an office workspace \cite{van1999provar}. Handy-1 is another adjustable table-mounted manipulator for ADLs such as eating, drinking, and washing applications \cite{topping2002overview}. The mounted robots were designed to perform various ADLs using a general-purpose manipulator. However, the limited workspaces of the robots restricts the range of available activities. Alternatively, researchers have introduced various wheelchair-mounted robotic arms (WMRAs). For meal assistance, Maheu et al. showed that people with  disabilities can feed themselves using a manually controlled JACO arm mounted on a wheelchair \cite{maheu2011evaluation}. Schroer et al. showed drinking assistance using a \mbox{7-DoF} KUKA arm \cite{schroer2015autonomous}. For object fetching, Kim et al. introduced the UCF-MANUS robot, consisting of a wheelchair-mounted manipulator and interface \cite{kim2014system}.  

\paragraph{\textbf{Mobile-base Robots}}
A mobile base can increase the workspace of a robot and the number of tasks it can perform. Hawkins et al. found that movement of a mobile manipulator's base was needed to provide assistance with a shaving task, since the PR2 that they used could not otherwise reach the relevant locations \cite{hawkins2014assistive}. In feeding, the fixed-base robot often requires the relocation of the robot or user by caregivers in the beginning or during the task. A fixed base restricts the scope of assistive tasks \cite{ari2015base}. Without mobility, robots are restricted to a narrow set of tasks and are unable to leave the immediate vicinity of the human to provide assistance elsewhere. Recent studies have introduced general-purpose mobile manipulators for various assistive robotic tasks, including shaving \cite{chen2013robots, hawkins2014assistive}, dressing \cite{yamazaki2014bottom, klee2015personalized, erickson2018tracking}, fetch-and-carry \cite{kargov2005development, Jain2009, graf2009robotic, ciocarlie2012mobile}, and guiding tasks \cite{jia2012design}. Our meal-assistance system has a mobile base that has the potential to enhance the quality of feeding assistance.

\begin{figure*}[ht]
\centering
  \includegraphics[width=0.8\textwidth]{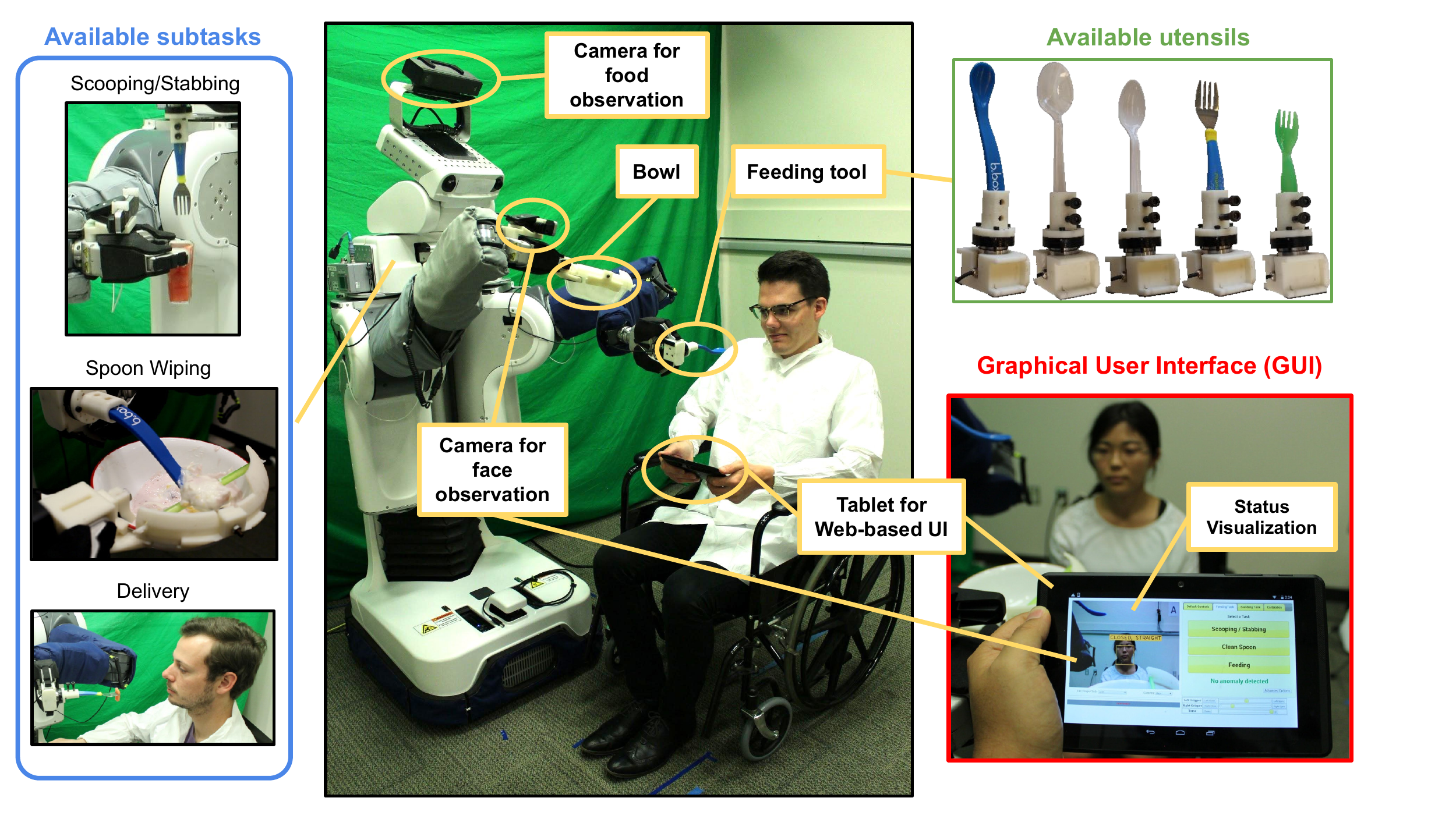}
\caption{The outline of our meal-assistance system using a general-purpose mobile manipulator. The system provides \textit{active feeding} assistance by delivering food from a bowl to a user's mouth. A user can select a preferred subtask via the graphical user interface. The robot then automatically estimates the location of food, scoops/stabs it depending on the current feeding tool, and places it inside a user's mouth. The images show able-bodied users.}
\label{fig:system_config}       
\end{figure*}
%

\subsection{Meal-Assistance Devices}
Researchers and companies have introduced various assistive devices for feeding. We focus on stabilizing/leveling handles, arm supports, and robots.

\paragraph{\textbf{Stabilizing \& Leveling Handle}} 
Tremor often causes spilling food or drink. Stabilizing handles can be used for damping tremor and smoothly transferring food on an attached spoon (e.g., Liftware Steady \cite{liftware}). 
Pathak et al. reported an electronically controlled handle could improve holding, transferring, and eating qualities during tremor-induced movements \cite{pathak2014noninvasive}. The self-stabilizing device is beneficial to people with Parkinson's disease. However, this type of device may not be suitable for people with severe tremor or other motor impairments.

The reduced hand and arm mobility due to weak muscles also cause difficulty leveling food on a utensil. Leveling handles such as Liftware Level \cite{liftware} can be used for aligning an attached utensil with gravity. However, this is for a certain range of people who can still lift their arm. 

\paragraph{\textbf{Arm Support}} 
Alternatively, arm support devices enable users to use their upper limb by supporting their weak arm, suppressing tremor, or expanding their limited arm movement. Several makers have developed wheelchair- or table-mounted arm supports: Neater \cite{neater} and Nelson \cite{nelson}. By moving a spoon attached on the device, users can scoop food and feed themselves while suppressing tremor or uncontrolled movements.
The devices can be powered or unpowered, but require users' manual movements using their upper limb. Thus, the comfort and efficiency of feeding depends on an individual's movement and would not be suitable for some users.

\paragraph{\textbf{Robot}} 
The use of meal-assistance robots is an alternative solution for people with various levels of motor impairments. Table \ref{tab:feeding_robots} shows a comparison result of features in currently available meal assistance robots that provide both food grasping and feeding functions. A number of commercially available solutions exist: Handy-1 \cite{topping2002overview}, Winsford feeder \cite{hermann1999powered}, My Spoon \cite{myspoon}, Bestic arm \cite{bestic}, Mealtime partner \cite{mealtime}, and Meal buddy \cite{meal_buddy}. These robots are designed for a particular purpose (i.e., meal assistance), often having a desk-mountable fixed base and a low DoF arm. 

A user can command a sequence of scooping-delivery motions via a joystick or a button. The robots follow predefined trajectories where food and mouth locations are hard coded. A recently released robot, Obi \cite{obi}, uses kinesthetic teaching from caregivers, but it still provides passive assistance in that it only moves to the specified location rather than adapting to the location of the user's mouth. To the best of our knowledge, there are no commercial feeding robots with adaptive movements, a powered mobile base, or human mouth tracking.

\begin{figure}[t]
\centering
\includegraphics[trim={0.0cm, 0.0cm, 0.0cm, 0.0cm}, clip, width=80mm]{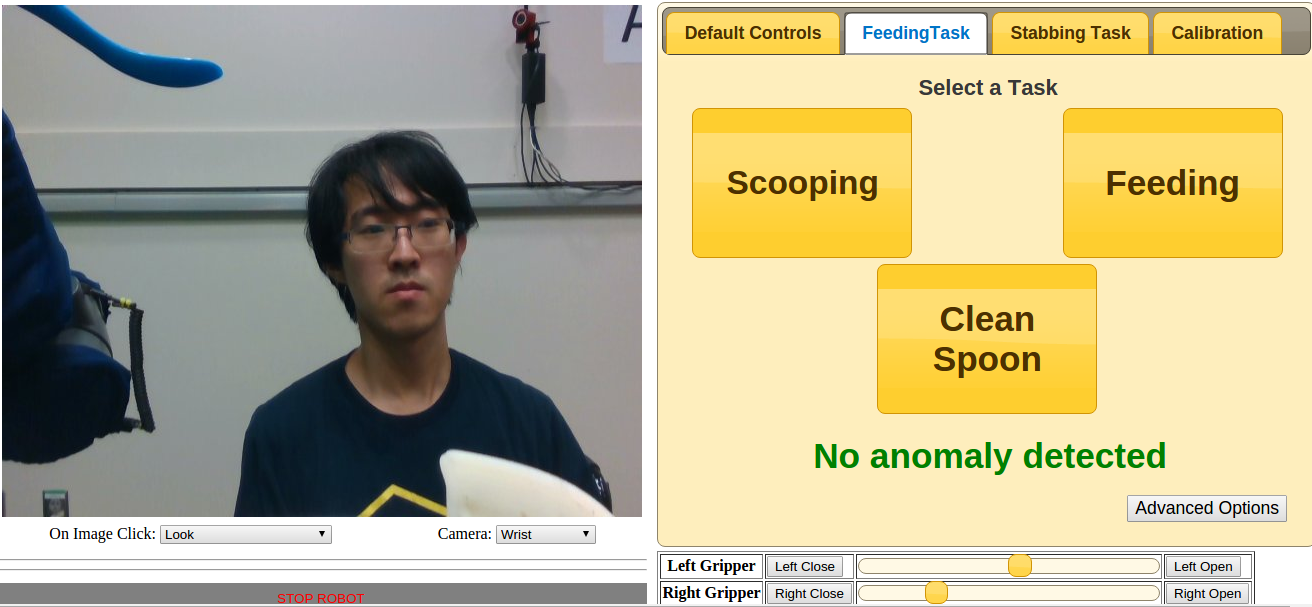}
\\
\includegraphics[trim={0.0cm, 0cm, 0cm, 0.0cm}, clip, width=80mm]{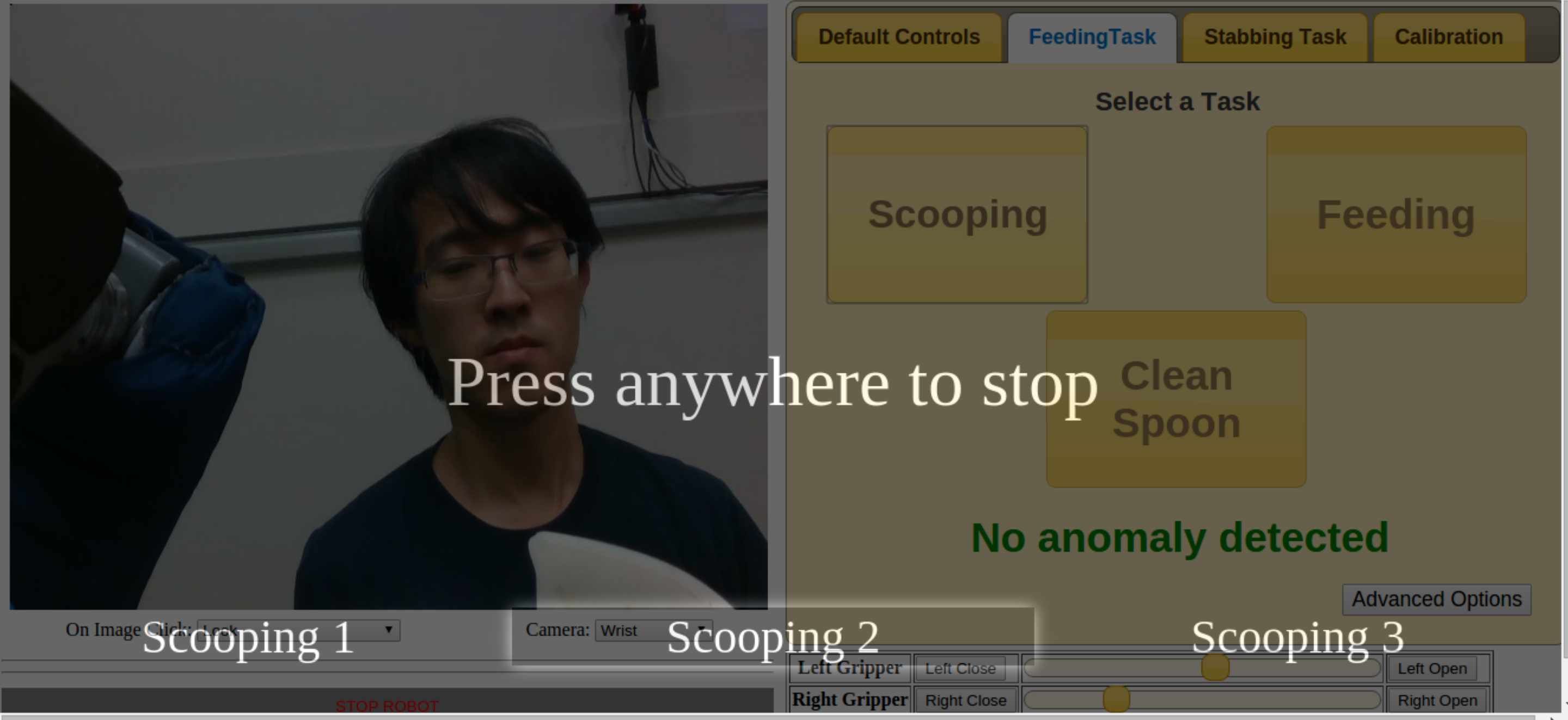}
\\
\includegraphics[trim={0.0cm, 0cm, 0cm, 0.0cm}, clip, width=80mm]{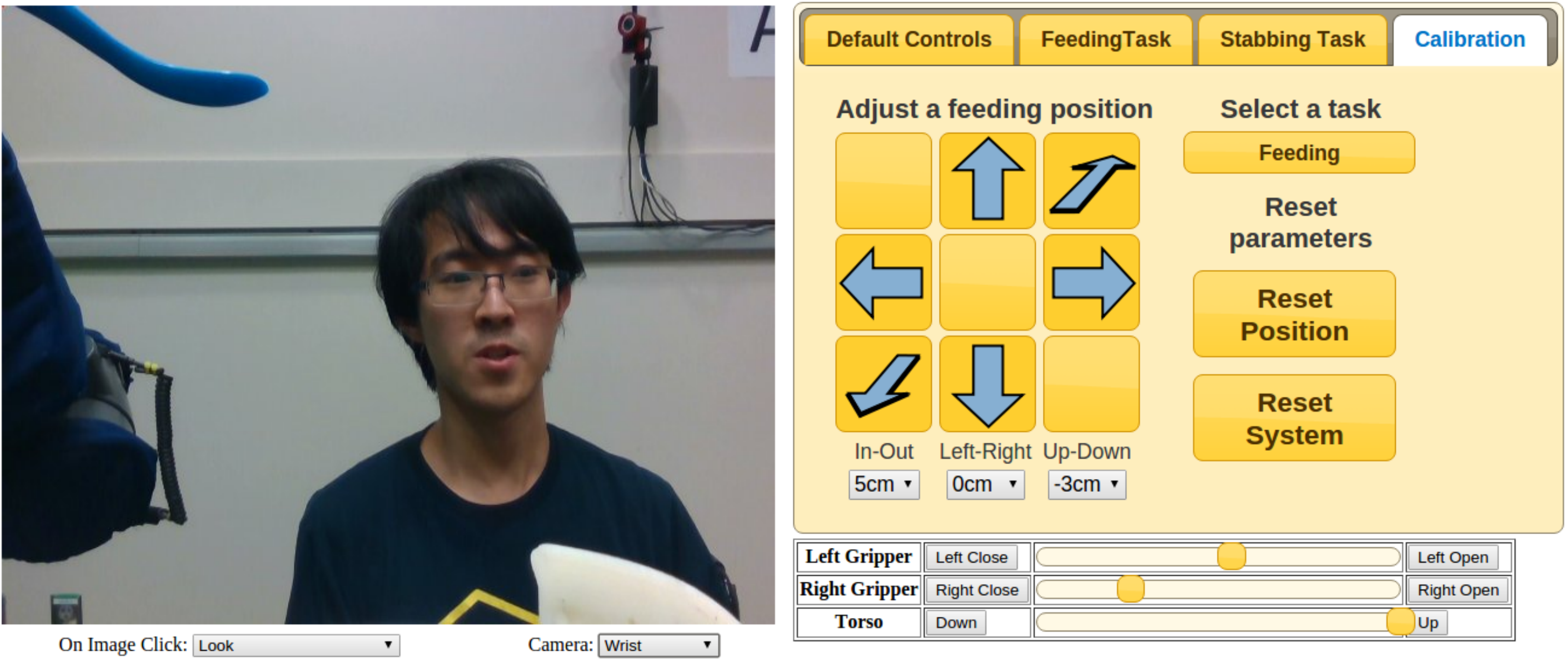}
\caption{Screen captures of our web-based graphical user interface (GUI). \textbf{Top} A main page when the system enters an idle state. A user can trigger a desired subtask by clicking one of three operation buttons on the right side. The GUI displays the visual outputs from a selected camera with the direction of the user's face and whether the mouth is open or closed. Our system uses this information to monitor anomalous task executions. \textbf{Middle} A full-screen stop button for quickly and easily stopping a running operation. We also provide task-state transition information on the bottom for users. \textbf{Bottom} A delivery-position calibration tab where users can add an offset to their preferred direction. This page also allows to execute a dry run without food. These images show able-bodied users only. }
\label{fig: gui}
\end{figure}

Researchers have introduced advanced meal-assistance systems with various functionalities \cite{naotunna2015meal}. Song and Kim designed a feeding robot with a specialized gripper for cultural food \cite{song2012novel}. Yamazaki and Masuda introduced a 5-DoF chopstick-equipped robot that provides various motions to pick foods of various characteristics \cite{yamazaki2012various,yamazaki2012autonomous}. 
The robot also allows each user to select a desired food-taking location via a graphical user interface (GUI). Recently, Admoni and Srinivasa introduced a gaze-based shared autonomy framework to predict a user's target piece of food and retrieve it \cite{admoni2017gaze}. Javdani et al. presented a shared-control teleoperation approach to orient the utensil \cite{javdani17}. Unlike these manual or semi autonomous systems, Kobayashi et al. introduced an automatic remnant food scooping method using a laser range finder \cite{kobayashi2016meal}. Similarly, our system selects a scooping location using an RGB-D camera and autonomously retrieves it.

In terms of delivering food, most robotic systems use \textit{passive feeding} executions in which a robot conveys food to a predefined location, typically in front of the user's mouth. These systems depend on the users' upper body movement to reach the food. Takahashi and Suzukawa, on the other hand, introduced an interface enabling a user with quadriplegia to manually adjust delivery locations \cite{takahashi2006easy}. Similar to our work, Schroer et al. proposed an adaptive drinking assistance robot that finds the user's mouth location with an external vision system \cite{schroer2015autonomous}. Recently, Candeias et al. introduced a visually-guided feeding system that also finds the user's mouth as well as checks food acquisition success using cameras \cite{candeias2018feeding}. We leverage such visual input to detect anomalies as well as the user's mouth. In addition, researchers have adapted feeding task movements to users' preferences by incrementally updating movement primitives \cite{canal2016personalization, 18iros-RhodesVeloso}.

\begin{figure}[h]
\centering
  \includegraphics[width=0.45
 \textwidth]{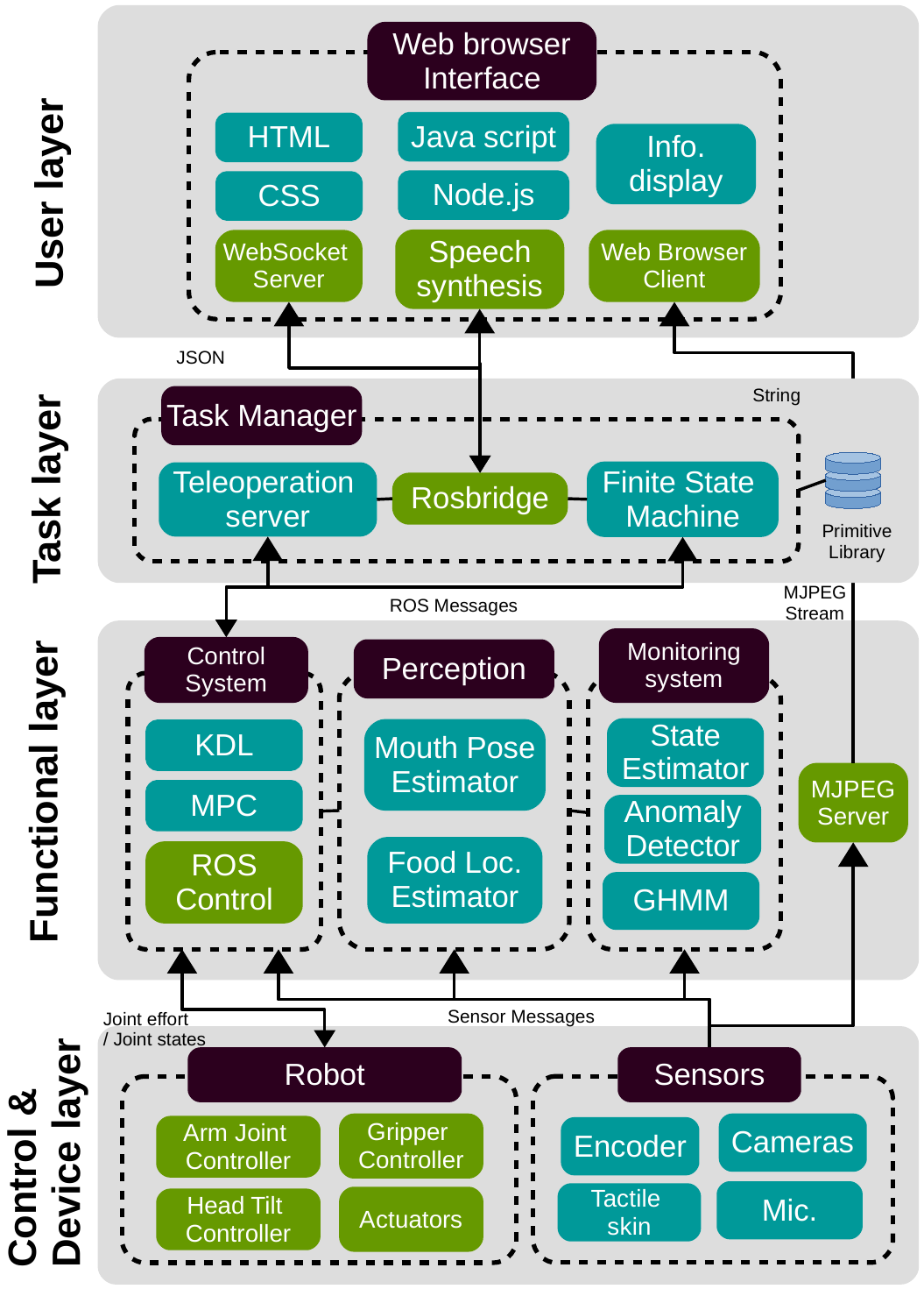}
\caption{The software architecture of our meal-assistance system. Users can command each subtask via the GUI. Through the robot operating system (ROS), our system executes and monitors the selected subtask.}
\label{fig:sw_system_config}       
\end{figure}

\section{Outline of System} \label{sec:3}
We briefly introduce our robot-assisted feeding system with high-level operating procedures and its components.

\begin{table*}[t] 
\caption{The summary of high-level operating procedures. Our meal-assistance system can perform three independent subtasks: scooping/stabbing, spoon wiping, and delivery. The multimodal information from the listed sensors enables the system to provide \textit{active} and \textit{safe} feeding assistance. The location of sensors are visualized in Fig.~\ref{fig: sensors}.}
\begin{center}
\begin{tabular}{p{2.3cm} p{10cm} p{3.5cm}}
\toprule
Subtasks & Details & Sensors \\
\midrule
    \multirow{2}{*}{\shortstack{Scooping / \\Stabbing}} & {The user clicks the Scooping button on the GUI described in Sec.~\ref{ssec: gui}. The robot then autonomously scoops a spoonful of yogurt from a random or visually selected location in the bowl using our visual food-location estimator described in Sec.~\ref{ssec: scooping} and \ref{ssec:est}.} & A head-mounted camera and joint encoders \\
\midrule
    Spoon wiping & {The user clicks the Wiping button if the scooping task results in excess food on the top or bottom of the spoon. The robot wipes off the surface of the spoon using the wiping bar on the bowl (see details in Sec.~\ref{ssec: wiping}).} & Joint encoders \\    
\midrule
    Delivery & {The user clicks the Feeding button on the GUI when an adequate amount of yogurt is present on the spoon. The user turns his or her head toward the camera on the robot's right wrist. The robot then estimates the pose of the user's mouth and delivers yogurt inside the mouth. It then pulls the spoon back from the mouth. (see details in Sec.~\ref{ssec: feeding}) To prevent potential injuries due to the robot's anomalous execution, the robot monitors the pattern of multi-modal sensory signals (see details in Sec.~\ref{ssec_safety}). We also run an optimization-based low-gain impedance controller (see details in Sec.~\ref{ssec: control}). } & A wrist-mounted camera, joint encoders, a microphone array, tactile skin sensors, a force-torque sensor, and current sensors \\
\bottomrule
\end{tabular}
\label{table: task_overview}
\end{center}
\end{table*}

\subsection{System Configuration}
Our robot-assisted feeding system hardware consists of a PR2 robot, tool holders, and additional sensors. Fig.~\ref{fig:system_config} shows an overview of the system. While providing visually-guided feeding assistance, the PR2 holds a bowl with its right end effector and a utensil in its left end effector. A user can command a preferred subtask via a graphical user interface (GUI) (see Fig.~\ref{fig: gui}). The PR2 is a 32-DoF mobile manipulator that consists of an omni-directional mobile base, a 1-DoF telescoping spine, and two 7-DOF back-drivable arms that are controlled by \SI{1}{\kHz} low-gain PID controllers. Its maximum payload and grip force are listed as \SI{1.8}{\kg} and \SI{80}{\N}, respectively. These are enough to firmly hold a bowl or a utensil during assistance.

The system can perform three independent subtasks: scooping/stabbing, spoon wiping, and delivery. For scooping, the system finds the highest food location in a bowl using a head-mounted RGB-D camera, Microsoft Kinect V2, and then scoops a spoonful of food from the bowl. For \textit{active feeding}, the system estimates the user's face and mouth pose using an Intel SR300 RGB-D camera mounted on top of the right wrist. While running the subtasks, the system runs a multimodal execution monitor to detect anomalous behaviors using 6 different sensors. We will discuss how to use these multimodal sensory signals to detect anomalous behaviors in Section~\ref{ssec: safety_sw}. We run all software components on top of the Robot Operating System (ROS) \cite{quigley2009ros}(see Fig.~\ref{fig:sw_system_config}). All our source code is available online\footnote{Code: \url{www.github.com/gt-ros-pkg/hrl-assistive}}.

\subsection{Operating Procedure}
In this section, we discuss the high-level operating procedure taken by the robot when a person with motor impairments wants to eat food, such as eating yogurt with a spoon. We assume the robot is placed at a location from which it can reach the user's mouth while holding a utensil and a bowl. We also assume the user can move and click a mouse pointer using a finger or a head (or eye) tracker. A user can then command a preferred subtask using the GUI. Table~\ref{table: task_overview} shows the summary of typical operations and a list of necessary sensors. During the operations, the user can stop and run the robot again whenever he or she wants using the interface. 

Our multimodal execution monitor runs in parallel with the scooping and delivery subtasks. When it detects an anomalous execution that largely differs from typical non-anomalous executions, the system pauses the current task execution and then moves the arm back to the starting initial pose of the current subtask. Note that the robot keeps the spoon level (parallel to the ground) to avoid food spills during the returning motions.

\section{System Components} \label{sec:4}
In this section, we introduce individual components of the feeding assistance system for people with disabilities.

\subsection{User Interface} \label{ssec: gui}
We introduce an improved version of a web-based GUI. Our previous work \cite{park2016towards, park2017class} introduced an earlier GUI. This GUI was based on a web-based GUI that transmits task commands and displays video from the camera for self-care tasks around a user's head \cite{hawkins2014assistive}. For the most recent version of the system, we improved the GUI to transmit calibration options, display task-relevant information, and collect feedback from users (see Fig.~\ref{fig: gui}).

To provide these functions, we created web pages using standard web technologies such as HTML5, CSS3, and JavaScript. We then enabled the pages to interact with ROS using a \textit{rosbridge} ROS Javascript library \cite{crick2017rosbridge}, which, using a \textit{Node.js} client, provides a JSON-based bidirectional communication interface between the web pages (i.e., client) and servers over Web-sockets. Our GUI also uses an HTTP-based Motion JPEG (MJPEG) server that transfers live video stream from the robot's cameras. We overlay the task state and its transition on top of the video stream. The overall data flow is visualized in the user and task layers shown in Fig.~\ref{fig:sw_system_config}.

The GUI is a device-agnostic interface that is used via a web browser. In our studies, people with disabilities have used the same interface via a tablet and a laptop. The interface consists of a live video screen that displays the video output from the head- or wrist-mounted cameras and a task tab that provides buttons or bars to command a subtask or adjust internal parameters of the system. 


In the feeding task tab, the user can select one of three subtask buttons: Scooping/Stabbing, Clean spoon, and Feeding (Delivery). The system then executes the selected subtask until finishing the subtask or receiving a stop command. The user can force the robot to stop at any moment by clicking a full-screen stop button (see Fig.~\ref{fig: gui} \textbf{Middle}) that appears during task executions. The stop command is treated as an anomalous event which triggers a corrective action following transition $T_A$ in the finite-state machine (FSM) described in Section \ref{ssec: fsm}. After task completion, the users may then enter feedback (i.e., \textit{success} or \textit{failure}). We used this user feedback to label the execution data for training and testing the execution monitor and tuning the performance of the system.

\begin{figure}[ht]
\centering
  \includegraphics[trim={0cm, 0cm, 0cm, 0cm}, clip, width=8cm]{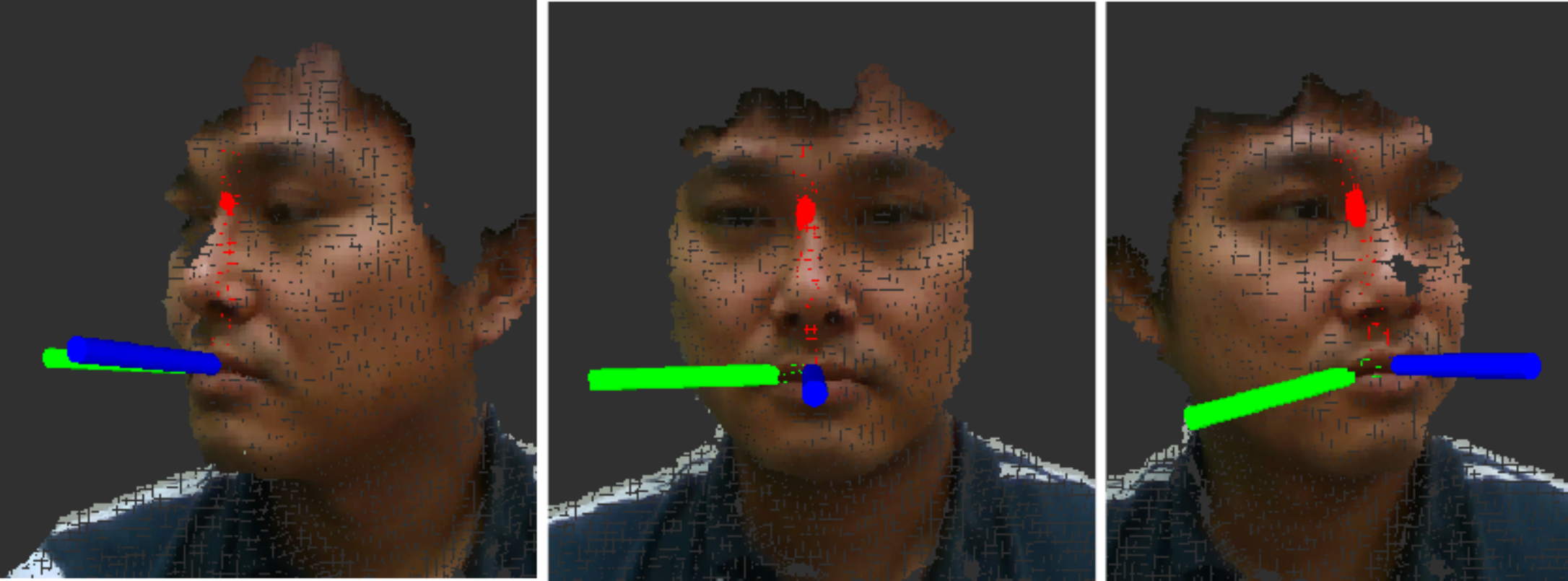}
\caption{Estimated mouth poses from our vision-based mouth-pose estimator.}
\label{fig:mouth_detector}
\end{figure}

In addition, users can adjust the delivery location where the robot places the utensil with food inside the user's mouth. By default, the robot places the tip of a utensil \SI{4}{cm} inside from the center of the estimated mouth plane (red-green plane in Fig.~\ref{fig:mouth_detector}). Fig.~\ref{fig: gui} \textbf{Right} shows the delivery calibration tab where a user can adjust their preferred delivery location with respect to the estimated pose of their mouth using 6 arrow buttons to add $\pm$\SI{1}{cm} offsets.

\subsection{Task Manager} \label{ssec: fsm}
The Task Manager, shown in the task layer of Fig.~\ref{fig:sw_system_config}, sequences action/observation states in three subtasks (i.e., scooping/stabbing, wiping, and delivery) using a finite-state machine (FSM).

\begin{figure}[t]
	\centering
    \includegraphics[width=87mm]{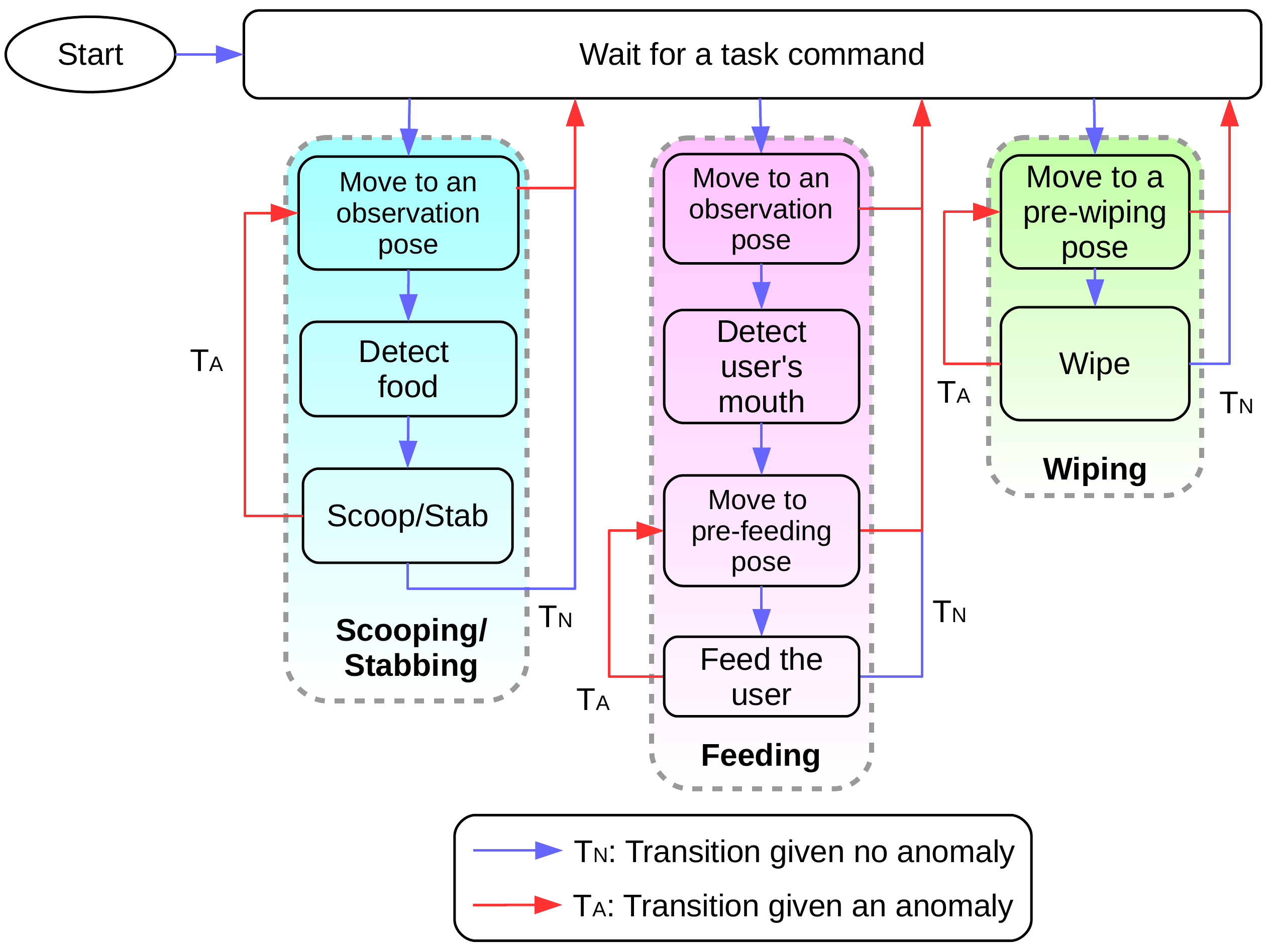}
	\caption{Finite-state machine for meal assistance. Each box represents a state, whereas arrows indicate state transitions.}
	\label{fig: motions}
\end{figure}

\paragraph{\textbf{Finite-state Machine}}
An FSM is a deterministic finite graph, which consists of states, transitions, and events. FSMs have frequently been used in robotic manipulation tasks due to their clear and easily implementable structure \cite{rosco_icra2013, Hebert2015robotsimian, tsagarakis2017walk}. Fig.~\ref{fig: motions} shows the system's FSM.

We incorporated two transition triggers to switch between states. The GUI enables the user to trigger the transitions $T_N$ and $T_A$, which are for non-anomalous and anomalous events, respectively. For example, the subtask buttons trigger $T_N$ and the full-screen stop button triggers $T_A$.
The multimodal execution monitor in Section \ref{ssec: safety_sw} can also trigger the transition for an \textit{anomaly}, $T_A$, when an anomaly is detected. Otherwise, a user can manually trigger it by clicking the full-screen stop button. The system then transitions to a state in which the robot halts or performs a corrective action. For instance, if a loud and unexpected sound is detected while feeding, $T_A$ is triggered and the robot will retract its arm to avoid harming the user. 

In the idle state, the robot waits for a user's command via the interface. Given each subtask command, the robot first initializes its configuration to a pre-defined configuration and then begins to execute the commanded subtask. For example, to perform the scooping subtask, the robot first initializes its configuration to a scooping pose and estimates the location of food in the bowl using a food location detector described in Section~\ref{sec:4}. After selecting a scooping location within the bowl, the robot then approaches the location to scoop a spoonful of soft food or stab a chunk of solid food. The user can then select to proceed to the delivery subtask if an adequate amount of food is present on the spoon. Otherwise, the user can command the robot to re-scoop/re-stab food or wipe off the spoon to remove excessive food.

\paragraph{\textbf{Scooping/Stabbing Subtask}} \label{ssec: scooping}
The scooping and stabbing subtasks aim to pick up and hold food. The system produces visually adapted scooping/stabbing motions using a sequence of motion primitives in which each primitive has a set of motion parameters, $\{ x_g, T_{\textit{Duration}}, \kappa \}$. $x_g$ is a goal pose ($\in \mathbb{R}^6$), which consists of position and orientation, $(p_g \in \mathbb{R}^3, q_g \in \mathbb{R}^3)$. $T_\textit{Duration}$ is the duration of the motion from its start pose to $x_g$ (e.g., \SI{3}{sec}). $\kappa$ is one of three typical motions: ``point to point" motion in joint space, ``point to point" motion in Cartesian space, and ``linear" end-effector motion in Cartesian space. Our control system in the functional layer generates a joint space / Cartesian space trajectory that satisfies the parameters.

The robot can also translate the goal pose with respect to the estimated food location. In this work, the robot can vary the target scooping location by visually estimating food locations within the bowl. For visual selection, we use a food-location estimator described in Section \ref{ssec: food_loc}. Note that we preregistered the size of the bowl so the system is able to estimate the scooping/stabbing area and restrict its motions to this region.
 
\begin{figure}[t]
	\centering
   \includegraphics[trim={0.2cm, 0cm, 0cm, 0cm}, clip, width=42mm]{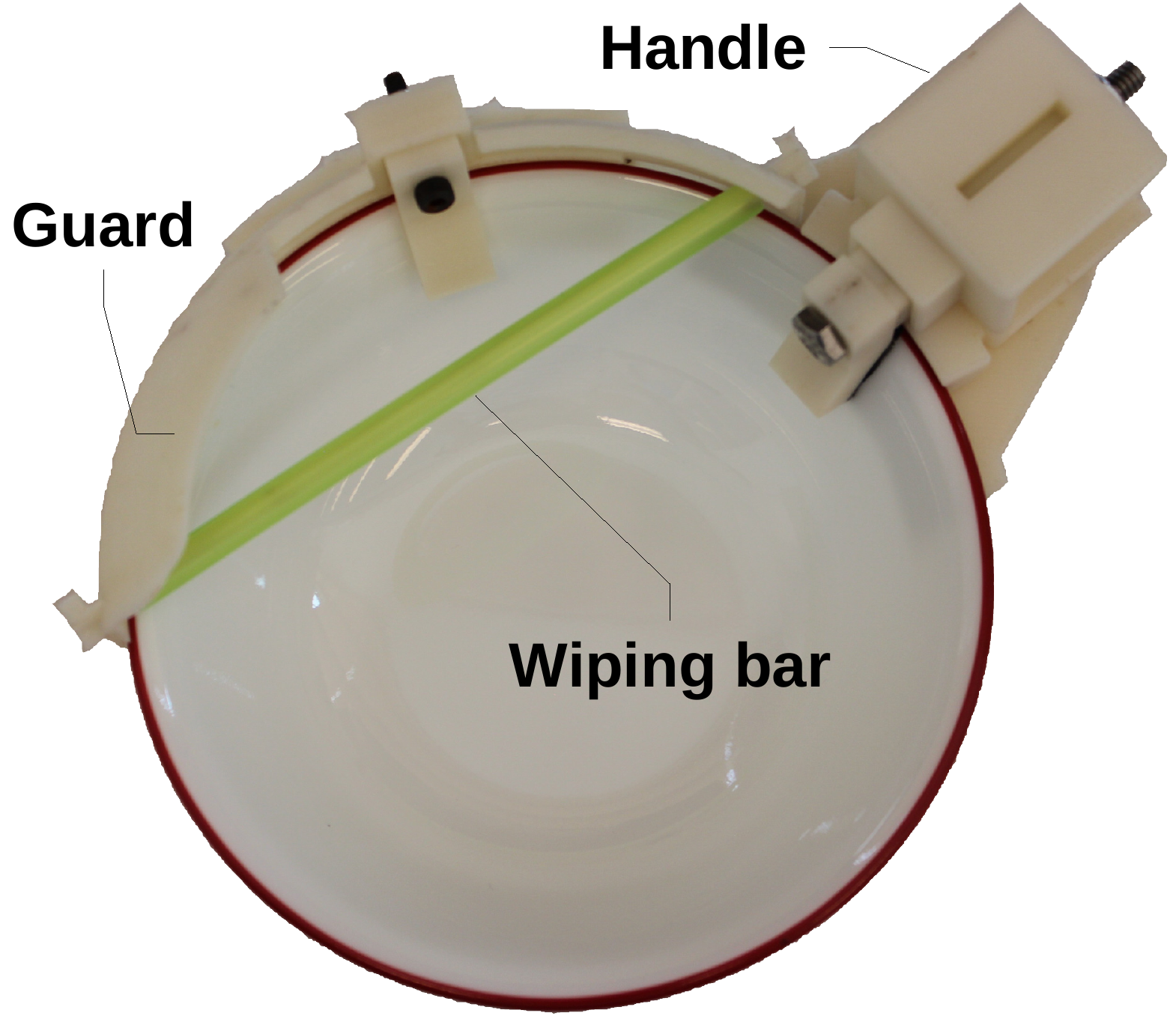}
   \includegraphics[trim=0.0cm 0.0cm 0.2cm 0.0cm, clip=true, width=42mm]{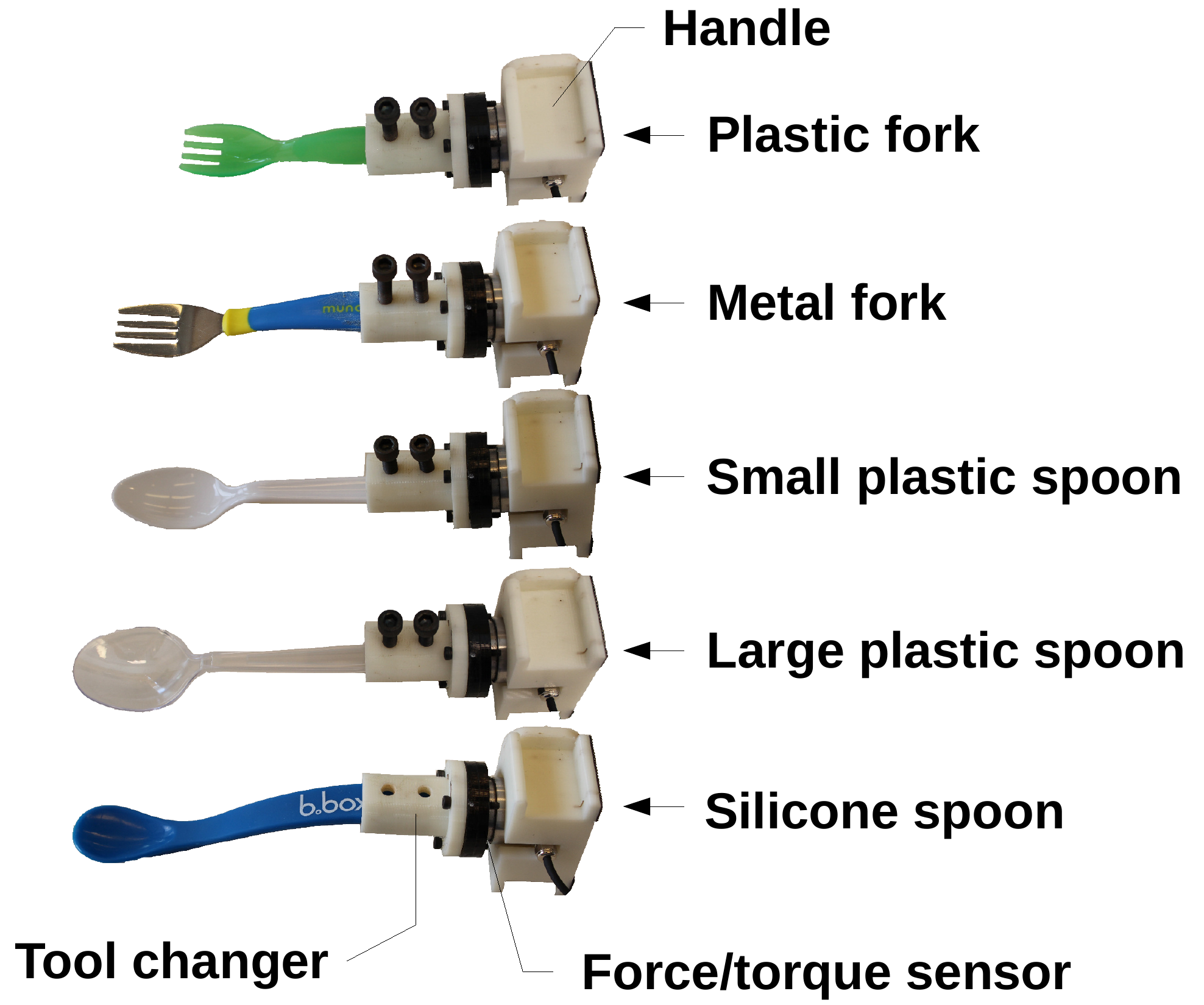}
	\caption{\textbf{Left:} A bowl with an attached handle and guard/wiping bar to avoid spilling food. \textbf{Right:} Five utensils: a flexible silicone spoon, small/large plastic spoons, a plastic fork, and a metal fork attached to a force-torque sensor.}
	\label{fig: feeding_tool}
\end{figure}

To account for various types of foods, a person, such as a caregiver, can mount a variety of utensils and bowls to the robot. Fig.~\ref{fig: feeding_tool} \textbf{Right} shows 5 representative utensils we used in our evaluation: a silicone spoon, small/large plastic spoons, a plastic fork, and a metal fork. A person can mount a preferred utensil into the 3D-printed tool-changer held by the robot and register transformation information from the tool to the utensil tip. 5 utensils are already preregistered in the system.
Fig.~\ref{fig: feeding_tool} \textbf{Left} shows a bowl our robot typically held during our evaluation. We also attached a handle to the bowl to enable a PR2 to easily grasp and hold it.

Food spilling can occur during scooping or stabbing due to excessive amounts of food and imperfect manipulation. To prevent spilling from the bowl, we mounted a 3D-printed bowl guard (see Fig.~\ref{fig: feeding_tool} \textbf{Left}). We also provided a cylindrical bar to wipe off excess food from the bottom of the spoon.

\begin{figure*}[t]
	\centering
   \includegraphics[trim={0cm, 0cm, 0cm, 0cm}, clip, height=50mm]{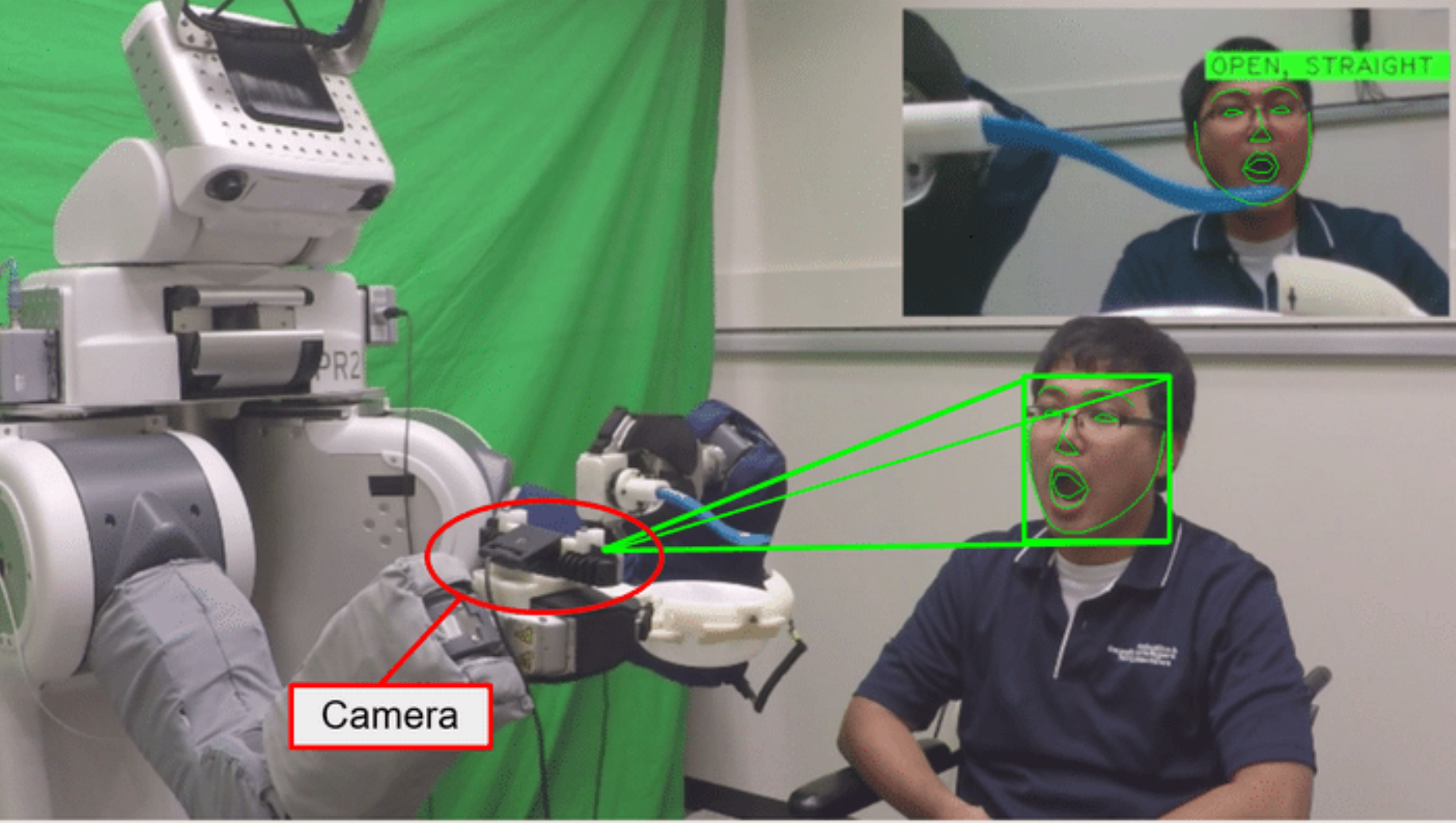}
   \quad
   \includegraphics[trim={0cm, 0cm, 0cm, 0cm}, clip, height=50mm]{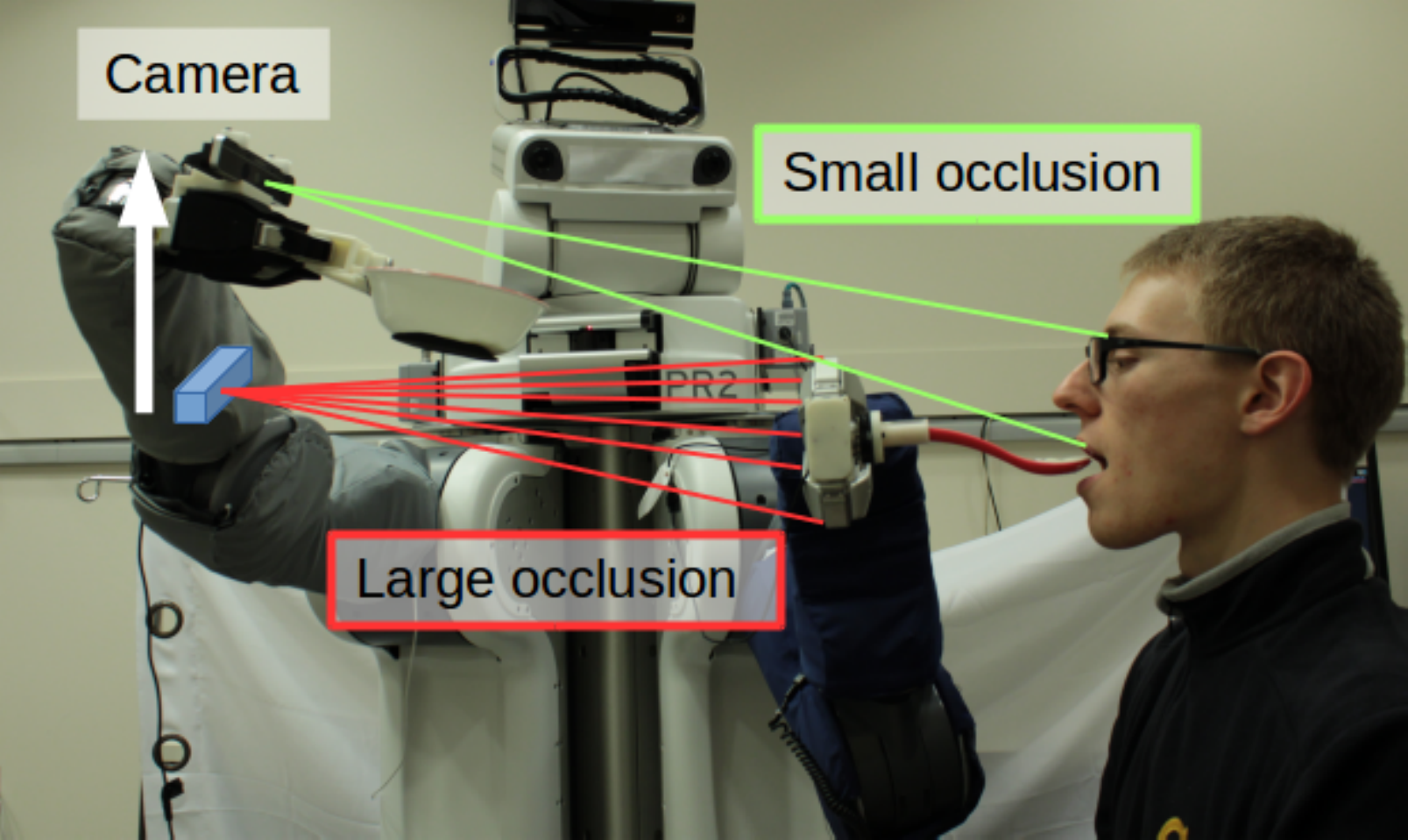}   
	\caption{Visualizations of facial landmarks and occlusions. Our system estimates the facial landmarks that may be occluded by the utensil or the robot's end effector. To reduce occlusions, the system lifts up the wrist-mounted camera during delivery.}
	\label{fig: camera_loc}
\end{figure*} 
\paragraph{\textbf{Wiping Subtask}} \label{ssec: wiping}
The scooping subtask often scoops excess food on the top of the spoon or some food at the bottom of the spoon. Both scenarios may result in food spills during feeding. A number of commercial robots use the edge of a bowl to clean excess food. Meal Buddy uses a wiping bar to wipe off excess food from the spoon and wipe drips from the bottom \cite{meal_buddy}. Our system addresses these issues by attaching 3D-printed food guard and wiping bar on the bowl (see Fig.~\ref{fig: feeding_tool} \textbf{Left}). The spill guard surrounding the bowl is \SI{3}{cm} high and it is used to block food spilling from the bowl while performing the scooping motions. The wiping bar is \SI{13.5}{cm} long and is used to remove excess food at the bottom of the spoon. The user can activate the wiping subtask by clicking the Clean Spoon button on the system's GUI. The robot drags the bottom surface of the spoon on the bar following a predefined linear trajectory. Note that the relative displacement between the right end effector and the bar is fixed due to the robot rigidly grasping the bowl.

\paragraph{\textbf{Feeding Subtask}} \label{ssec: feeding}
The feeding subtask aims to provide easily accessible and safe meal-assistance to a wide range of users, such as people with quadriplegia. Unlike conventional \textit{passive feeding} systems in the literature, our \textit{active feeding} system does not require a user to have upper body/limb movement in order to eat food off the utensil. In comparison, our system automatically delivers food inside the user's mouth. 

To put food inside the mouth, the robot uses a 3D mouth-pose estimator along with a wrist-mounted RGB-D camera that allows the robot to observe the user's face, as detailed in Section \ref{ssec: mouth_pose}. After estimating the mouth pose, the system selects a delivery position inside the mouth with a predefined or user-selected offset, as discussed in Section~\ref{ssec: gui}. After the first successful feeding attempt, the robot stores and re-uses the estimated mouth pose to shorten consecutive feeding times. Our system also provides a button to re-estimate the mouth pose when the user wants (see Fig.~\ref{fig: gui} \textbf{Right}).

Our system does not require pose teaching by users or caregivers. The system instead uses a set of predefined end-effector trajectories, linearly interpolating predefined end-effector poses with respect to a predefined mouth coordinate frame visualized in Fig.~\ref{fig:mouth_detector}. After estimating a person's mouth pose, the system transforms the trajectories from the mouth coordinate system to the world coordinate system. For some utensils, such as the large plastic spoon, we observed that during our pilot studies some users experienced difficulty eating food off the spoon due to its deep concave shape. As a result, our system also performs spoon tilting motions once the spoon has entered a user's mouth to make eating from such utensils easier.

For safety, the system observes the user's face and the utensil during feeding. However, the robot may not fully observe the face due to occlusion by the other end effector and the utensil it holds. To reduce some of these occlusions, our system lifts up the camera while feeding, which enables the landmark estimator, described in Sec.~\ref{ssec:est}, to more accurately predict facial points (see Fig.~\ref{fig: camera_loc}).

\subsection{Control} \label{ssec: control}
Our manipulation module, shown in the functional layer of Fig.~\ref{fig:sw_system_config}, executes planned motion primitives by generating joint torque commands $\tau \in \mathbb{R}^n$, where $n$ is the number of joints. To avoid having a stiff arm make contact with the human body, we use a \SI{50}{Hz} mid-level model predictive controller (MPC) \cite{jain2013reaching} followed by a \SI{1}{kHz} low-level PID controller with low gains. The MPC outputs a desired change of joint angles $\Delta \theta^*$ that minimizes a quadratic objective function including contact force and position error.

We improved the original MPC by adding orientation control in the objective function: 
\begin{align}
    \Delta \theta^* = \argmin{\Delta \theta} &\begin{Vmatrix} { \Delta p_d \choose \Delta q_d } - J_{ee} \left( K + \sum\limits_{i=1}^m J_{c_i}^T K_{c_i} J_{c_i} \right)^{-1} K \Delta \theta
    \end{Vmatrix}^2  \label{eq_mpc}\\ 
    \text{subject to} &\ \Delta \theta_{min} \leq \Delta \theta \leq \Delta \theta_{max} \nonumber \\
    &\ \theta_{d} = \theta + \Delta \theta , \nonumber
\end{align}
where $\Delta \theta$, $\Delta p_d$, and $\Delta q_d$ are the changes of joint angles, end-effector position, and end-effector orientation changes, respectively. We obtain the orientation changes via spherical linear interpolation planning for an end-effector trajectory. $J_{ee} \in \mathbb{R}^{6 \times n}$, $K \in \mathbb{R}^{n \times n}$, $K_{c_i} \in \mathbb{R}^{3 \times 3}$, and $J_{c_i} \in \mathbb{R}^{3 \times n}$ are the Jacobian matrix at the end effector, the joint stiffness matrix, the contact stiffness matrices, and the Jacobian matrices at the $i$th contact point on the robot's skin sensor, respectively. Further details are available in \cite{jain2013reaching}.

In this work, we set $K_{c_i}$ to zero to achieve consistent motions. This prevented the possibility of the arm's motion being changed due to a false positive detection of contact by the tactile sensing sleeve. This results in the following simplified form of Eq.~\ref{eq_mpc}.
\begin{align}
    \Delta \theta^* = \argmin{\Delta \theta} &\begin{Vmatrix} { \Delta p_d \choose \Delta q_d } - J_{ee} \Delta \theta
    \end{Vmatrix}^2  \\ 
    \text{subject to} &\ \Delta \theta_{min} \leq \Delta \theta \leq \Delta \theta_{max} \nonumber \\
    &\ \theta_{d} = \theta + \Delta \theta. \nonumber
\end{align}

The low-level PID controller enables the robot to track the desired joint angle $\theta_{d} \in \mathbb{R}^n$:
\begin{align}
    \tau = K (\theta_{d} - \theta) - D \dot{\theta} + \hat{\tau}_g,
\end{align}
where $D$, $\theta$, $\dot{\theta}$, and $\tau_g \in \mathbb{R}^n$ are the damping matrix, the current joint angles, the current joint velocity, and a gravity-compensation torque vector, respectively. For safety, we use empirically determined low gain matrices, $K$ and $D$.
The diagonal entries of $K$ and $D$ listed in order from the most proximal to the most distal joint of an arm were  $[90,80,20,22,12,27.5,20]$~$\SI{}{\newton.\metre/\radian }$ and $[10,10,2,1,1,2,2]$~$\SI{}{\newton.\metre.\sec/\radian }$, respectively. These values result in low stiffness of the arms.

\subsection{Estimators}\label{ssec:est}
Our perception component, shown in the functional layer of Fig.~\ref{fig:sw_system_config}, consists of two independent modules that estimate the food and user mouth states.

\paragraph{\textbf{A Food-location Estimator}} \label{ssec: food_loc}
The food-location estimator finds a location where the robot can scoop or stab food from a bowl. This estimator mitigates the chance of the robot making several scooping attempts without getting any food. An example of a different strategy it provided by the commercial feeding robot, iEAT \cite{ieat}, which rotates its food plate to adjust the scooping point using fixed, pre-defined motions. In contrast, we address this issue with a visually-guided motion based on the output of a vision-based estimator.

\begin{figure}[t]
\centering
  \includegraphics[trim={0cm, 0cm, 0cm, 0cm}, clip, width=8.8cm]{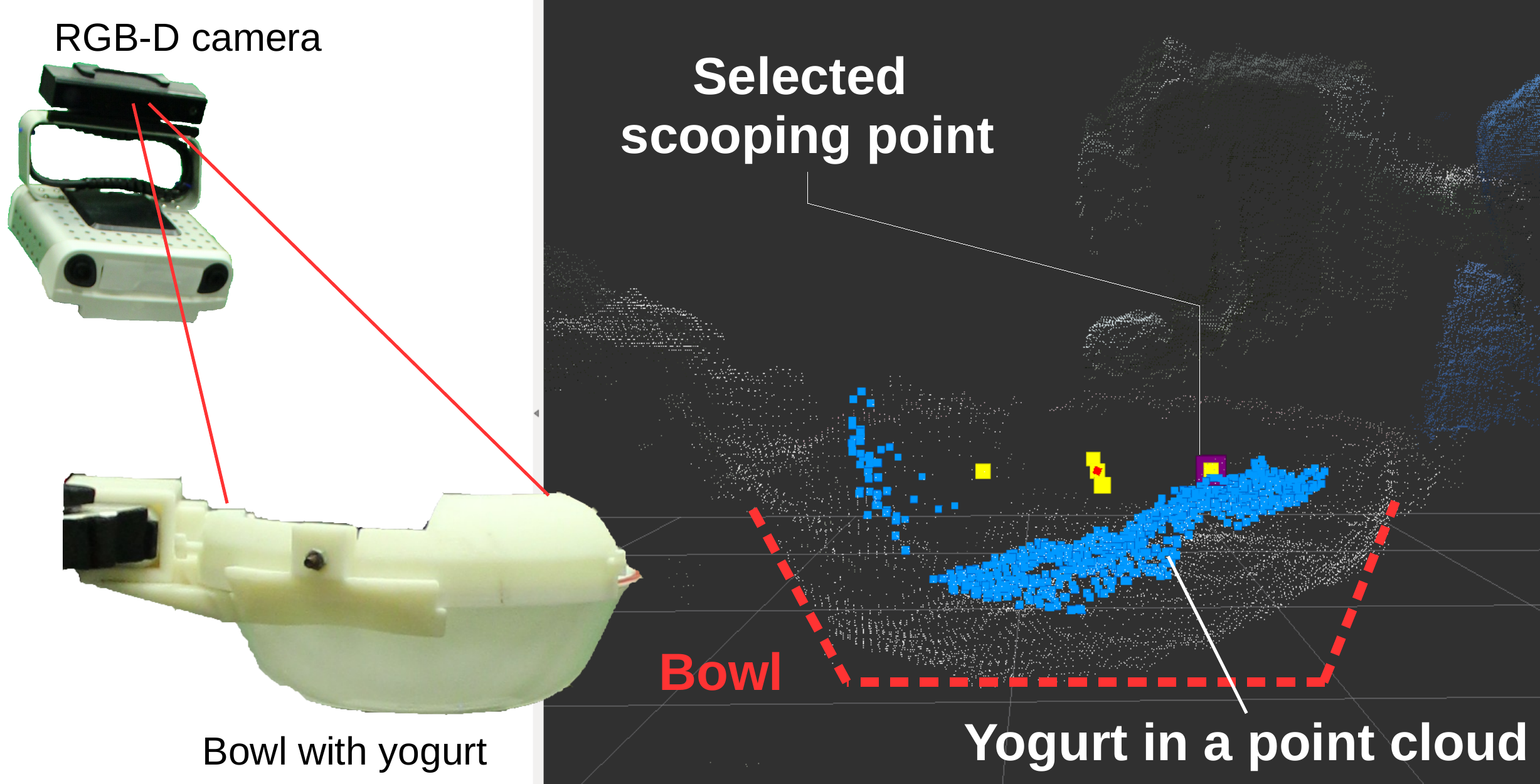}
\caption{The food-location estimator finds a location with relatively more food using the point cloud. Blue points indicate the top surface of food inside the bowl. Yellow points represent 5 potential scooping/stabbing locations and the selected location is shown in red.}
\label{fig:food_locater}
\end{figure}
%
%
The food-location estimator attempts to find the best scooping or stabbing location $s^* \in \mathbb{R}^3$ given a food point cloud, $\Upsilon$, from the head-mounted RGB-D camera (see Fig.~\ref{fig:food_locater}) and the geometry of the bowl, such as its center and diameter, as prior knowledge. We discretize the space of locations into five locations, $\mathcal{S} = \{s_1, s_2, s_3, s_4, s_5 \}$, displayed as yellow markers in Fig.~\ref{fig:food_locater}. To exclude an irrelevant part of the point cloud like the curvature of the bowl, we define a binary decision function, $\Psi: x \rightarrow \{0,1\}$, that returns $0$ if a point location $x \in \mathbb{R}^3$ is outside an ellipsoidal area in the bowl. 

We then compute the score of the scooping or stabbing location as a weighted sum of food point counts. The weight is the multivariate Gaussian probability density score centered at each location $s_i$ with a sample co-variance matrix of the food point cloud $\bar{\Sigma}(\Upsilon)$. We then find a location with the highest score:
\begin{align}
s^* &= \argmax{s_i \in \mathcal{S}} \sum\limits_{x}\frac{\exp\left(-\frac 1 2 ({x}-{s_i})^\mathrm{T}{\bar{\Sigma}(\Upsilon)}^{-1}({x}-{s_i})\right)}{\sqrt{(2\pi)^k|\bar{\Sigma}(\Upsilon)|}} \Psi(x).
\end{align}

\begin{figure}[t]
\centering
  \includegraphics[trim={0cm, 0cm, 0cm, 0cm}, clip, width=5cm]{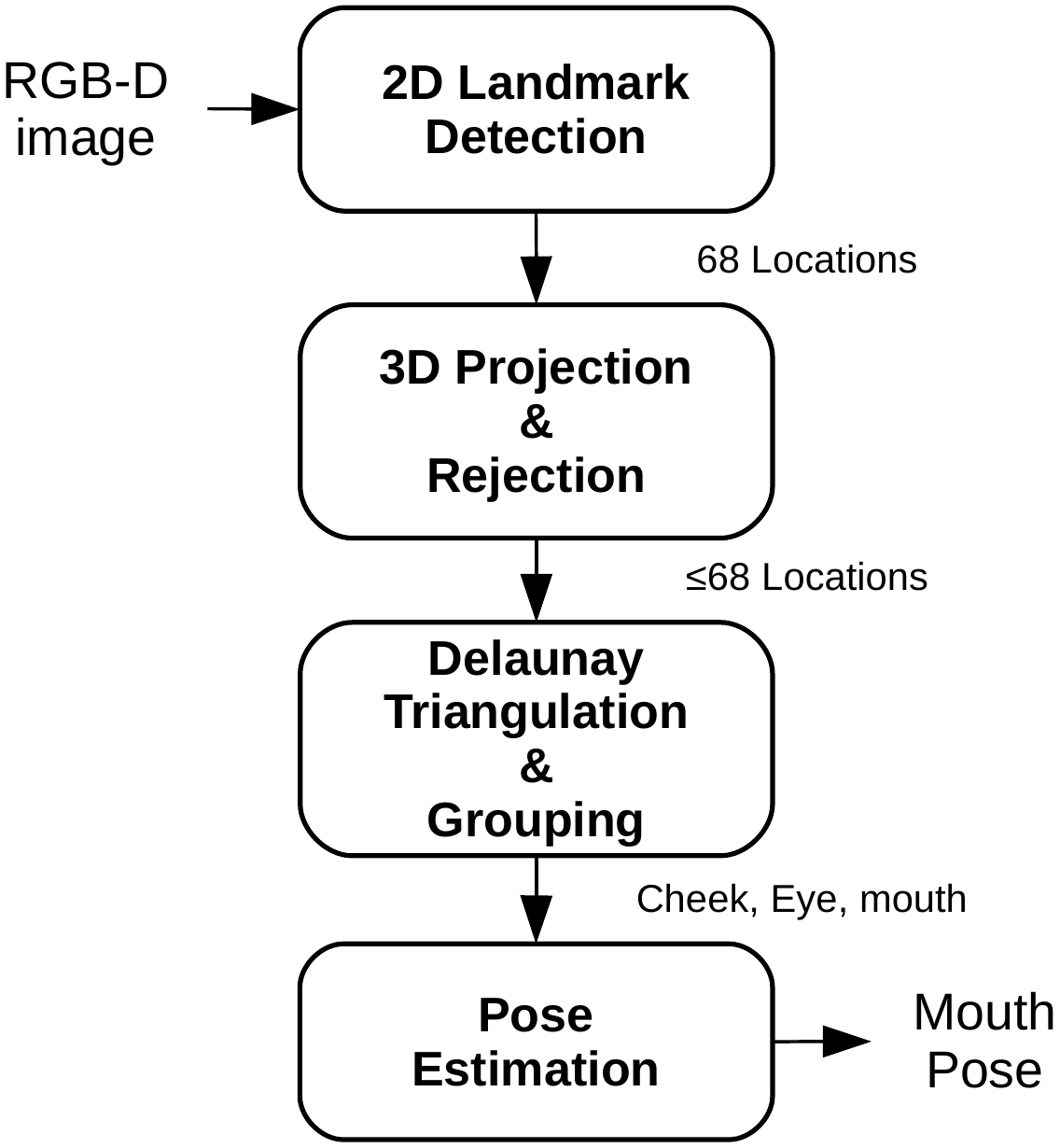}
\caption{Estimation framework for mouth-pose. We use RGB-D input to estimate the 6D pose of a user's mouth. }
\label{fig:estimation}
\end{figure}

\paragraph{\textbf{A Mouth-pose Estimator}} \label{ssec: mouth_pose}
Next, we describe a mouth-pose estimator with implementation details. The estimation of a user's mouth pose plays an important role in enabling the robot to provide \textit{active feeding} assistance to diverse users. In our previous work \cite{park2016towards}, the robot first estimated the location of an ARTag attached to the user's forehead and then  estimated the location of the user's mouth with a predefined rigid transform from the ARTag's pose. In this work, our mouth-pose estimator does not require an ARTag and rigid transform. Instead, it directly estimates the location of the user's mouth from an RGB-D image.

As seen in Fig.~\ref{fig:estimation}, 
our estimator first uses input from the wrist-mounted camera to extract facial landmarks \cite{SAGONAS20163}, key points of interest to localize facial regions such as the mouth, nose, left eye, right eye, and jaw (see Fig.~\ref{fig: camera_loc}). This process uses the frontal face detector in the open source \textit{dlib} library \cite{dlib09,Kazemi2014} in which the method localizes a user's face from an RGB image and detects 68 landmarks from a frontal face using the Histogram of Oriented Gradients (HOG) feature combined with a linear classifier and sliding window detector. The process is made robust to light variations by normalizing each window of the histogram. 

Our algorithm converts these 2D locations to 3D points by projecting them onto a depth image. This process can produce large errors due to the noise in the depth data and poor time-synchronization between the RGB and depth images. Thus, our algorithm rejects landmarks that 1) have large 3D distances from frontal-face reference landmarks defined with respect to the current mouth coordinate frame and 2) have large displacements ($\leq$\SI{5}{\cm}) over time ($\leq$\SI{10}{Hz}). To obtain the reference landmarks, the system registers a user's face during the first feeding execution by collecting a set of 20 reference landmarks. Then, following the model-based face localization method in \cite{Garcia2008}, our algorithm also rejects landmarks that differ largely from pre-modeled landmarks located near the eyes as false positives. After this rejection process, our algorithm computes a Delaunay triangulation of the landmarks to approximate the surface of the face \cite{itseez2015opencv} and then groups the landmarks in the model into three groups: cheek, eye, and mouth. Finally, the estimator determines the position and orientation of the mouth at the center of the mouth group and perpendicular to the plane defined by the center points of the three groups.


\subsection{Safety System}\label{ssec_safety}
Safety is an important consideration for feeding assistance. Ideally, robot-assisted feeding would be safe and effective without close supervision by a caregiver. This could enable people with disabilities to have greater independence and reduce caregiver burden. When compared to a simple specialized assistive feeding device, the greater autonomy and complexity of our \textit{active feeding} system increases the chance of an error occurring. In this section, we describe aspects of our system that we have designed to improve safety.

\begin{figure}[t]
\centering
  \includegraphics[trim={0cm, 0cm, 0cm, 0cm}, clip, width=8.8cm]{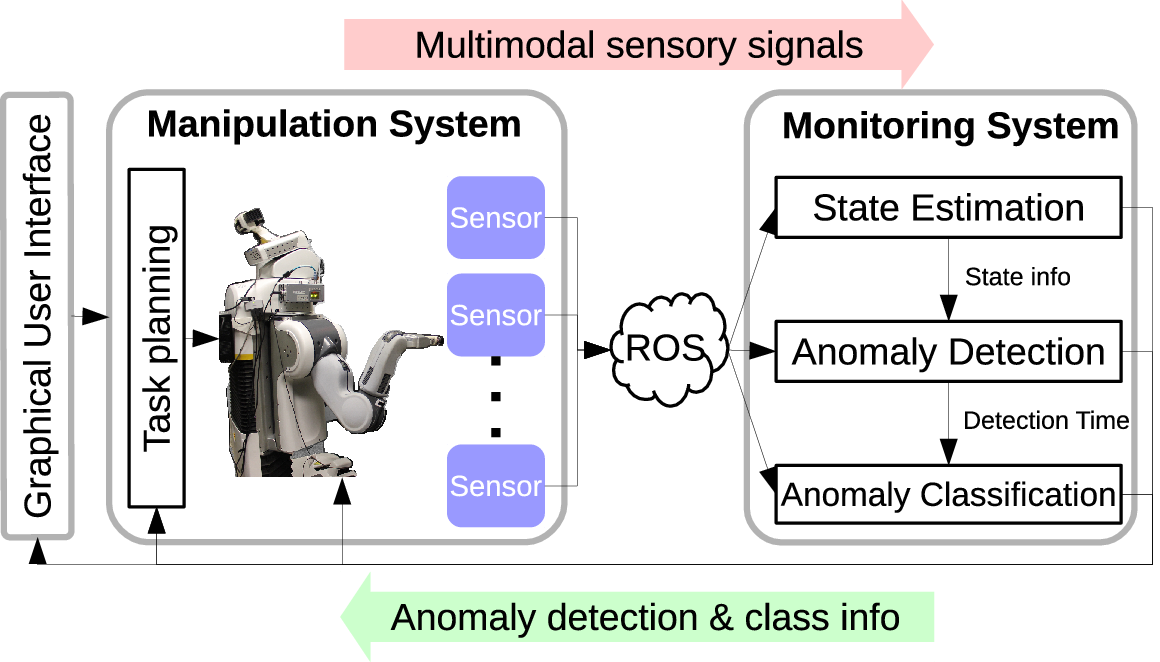}
\caption{Overview of the multimodal execution monitor, which estimates the task execution state, detect an anomaly~\cite{park2018detection}, and classify the cause of the anomaly~\cite{park2017class} for safe feeding assistance.}
\label{fig:ad}
\end{figure} 
\paragraph{\textbf{Hardware}} \label{ssec: hw}
The PR2 arms are backdrivable and controlled by a \SI{1}{kHz} low-gain PID controller to reduce the forces applied in the event of a collision. The PR2 also provides an run stop button that a user or a caregiver can press to cut off power to the PR2's motors. In the event that the system loses powers, the PR2's passive spring counterbalance system helps to keep the arms from descending rapidly due to gravity \cite{garage2012pr2}.

\paragraph{\textbf{Interface}} \label{ssec: safety_sw}
Our GUI provides a full-screen stop button (see Fig.~\ref{fig: gui} \textbf{Middle}) for people with motor impairments to conveniently and quickly cancel the current subtask and stop the robot's current motion. During task executions, the button expands to the entire web browser and, if the user clicks anywhere on the browser, the \textit{rosbridge} server for the GUI sends a stop command to the system. The command then triggers the $T_A$ transition on the FSM (Fig.~\ref{fig: motions}), which instructs the robot to return to the initial pose of the current subtask.

\paragraph{\textbf{Multimodal Execution Monitor}} \label{ssec: EM}
 We have previously presented our research on a multimodal execution monitor \cite{park2018detection, park2017class} that enables a robot to detect and classify anomalies during robot-assisted feeding (see Fig.~\ref{fig:ad}). By detecting and classifying anomalies, a robot has the potential to operate more safely, such as by stopping and alerting a caregiver if an anomaly has occurred. The system we describe in this paper includes a version of our multimodal execution monitor that detects and classifies anomalies. However, we adjusted the detection threshold to reduce the sensitivity of the anomaly detector during our study with 8 people with motor impairments. We made this adjustment to avoid false detections of anomalies during the study, since anomaly detection was not the focus of this study and an experimenter was always present and ready to press a button to stop the robot.

\begin{figure}[t]
	\centering
    \includegraphics[trim={0cm, 0cm, 3cm, 0cm}, clip, width=85mm]{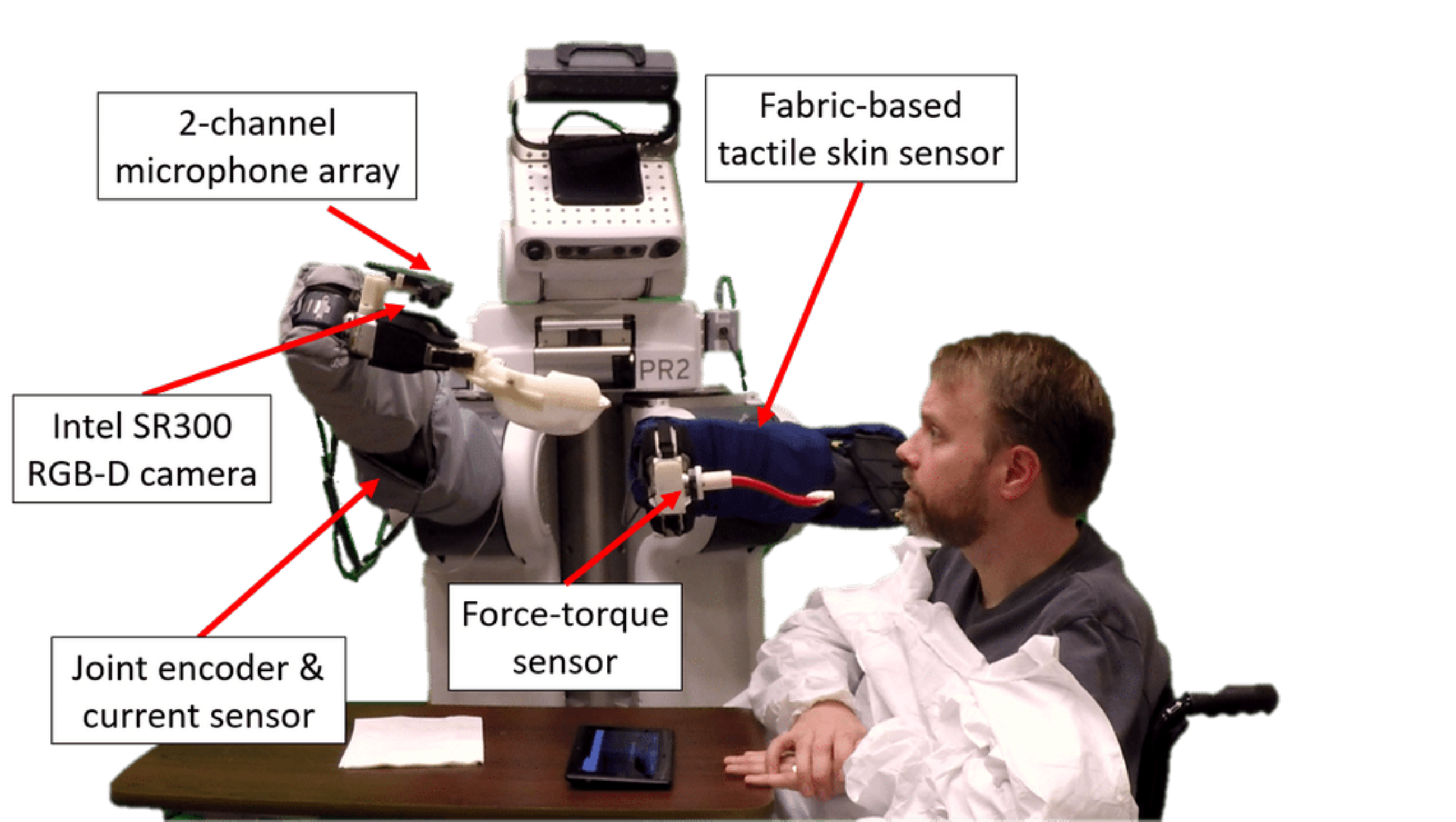}
	\caption{The various sensors we used to monitor task executions and anomalies. This figure shows a meal-assistance demonstration for a person with a self-feeding disability due to motor impairments.}
	\label{fig: sensors}
\end{figure} 
%

After detecting an anomaly, the monitoring system forces the system to stop and move its arm back to the initial pose for safety, as defined by the FSM. An anomaly classifier then fuses available features and estimates the most probable type and cause of 12 common anomalies. 
The features come from the multiple sensors mounted on the PR2: SR300 RGB-D camera with a two-channel microphone, joint encoders \& current sensors, fabric-based tactile skin sensors, and an ATI force/torque sensor (see Fig.~\ref{fig: sensors}). Our classifier consists of a multilayer perceptron (MLP) which takes as input temporal features extracted from HMMs and convolutional features from a convolutional neural network (CNN). The resulting information is sent to the robot's system to correct the current execution, plan a recovery strategy, or potentially improve the robotic assistance system. The classifier can enable a robot to detect, classify, and respond appropriately to common anomalies for effective and safer assistance. Further details are described in \cite{park2017class}.

\begin{figure}[t]
	\centering
 	  \includegraphics[width=80mm]{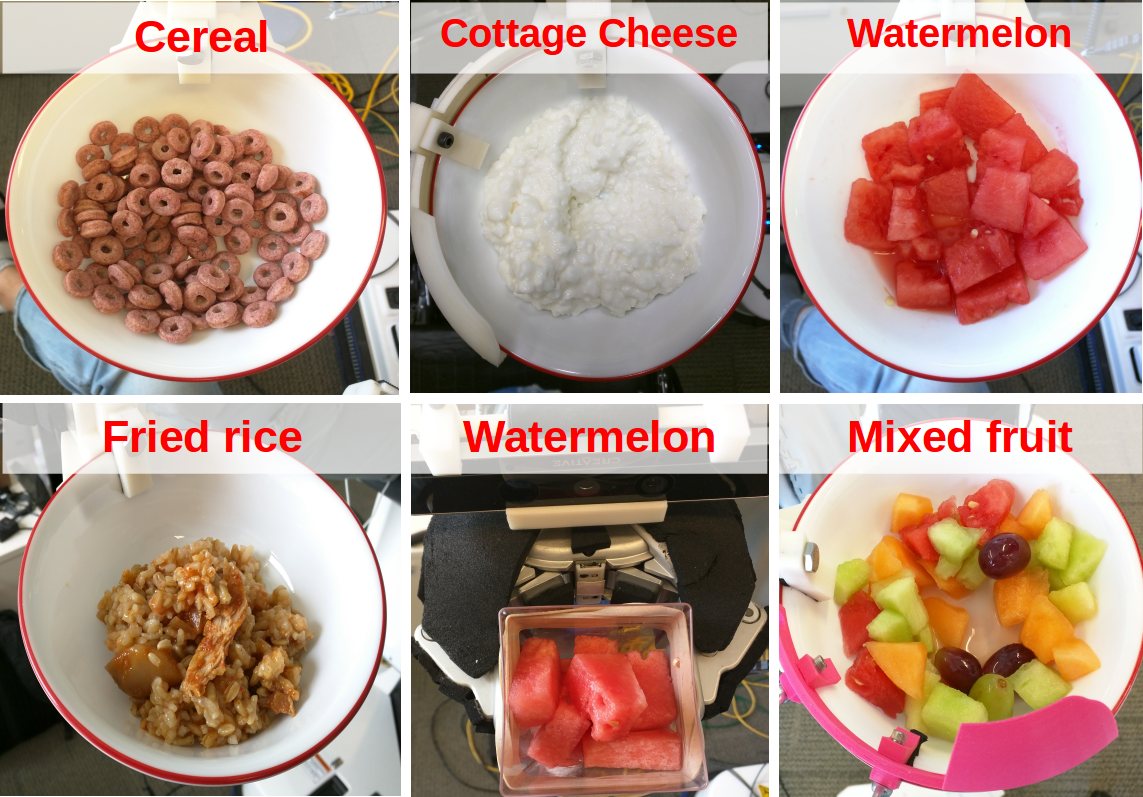}
	\caption{Examples of foods we used during the long-term evaluation.}
	\label{fig: self_study_food}
\end{figure}

\section{Experimental Setup} \label{sec:5}
Prior to our main study with people with motor impairments, we conducted an evaluation of the meal-assistance system with 9 able-bodied participants ($N=9$), the first author performed a long-term self evaluation, and we performed a three-day evaluation with Henry Evans in his home in California, USA. Through these evaluations we developed and confirmed the usability and safety of the \textit{active feeding} system. We then evaluated the system with 8 people with motor impairments that restricted their self-feeding ability ($N=8$) in the Healthcare Robotics Lab at Georgia Tech, Georgia, USA.  We conducted all of these evaluations with approval from the Georgia Tech Institutional Review Board (IRB).

Participants controlled the robot to scoop/stab food and feed themselves through our web-enabled GUI. For all of the studies other than the in-home evaluation with Henry Evans, participants accessed the interface through a \SI{7}{inch} Google Tango tablet with a Chrome browser, while sitting next to the PR2 robot. Henry used his personal laptop, headtracker, and mouse to access the interface with a Chrome brower. Before starting this evaluation, we briefly trained the participants to use the meal-assistance system. As part of this training, they practiced using the system three times. Each practice run took about one minute. They freely controlled the robot to wipe off the bottom of the spoon before they ate food. For safety during our evaluations, an experimenter was always present with a run stop button. 


\begin{figure}[t]
	\centering
    \includegraphics[trim={0cm, 0cm, 0cm, 0cm}, clip, width=85mm]{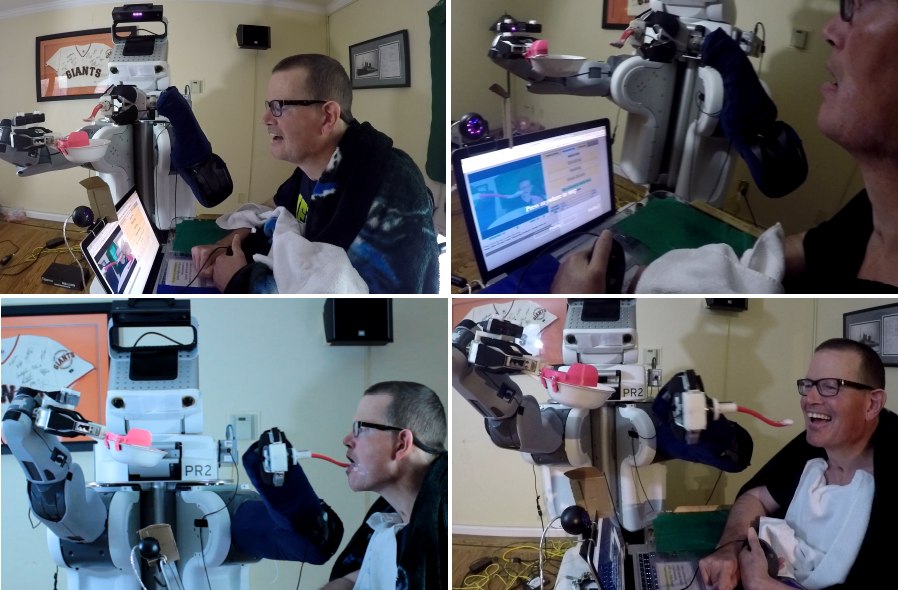}    
	\caption{The robot-assistive feeding system was deployed and tested in the home of Henry Evans, a person with severe motor impairments. Henry used the system for two sessions per day under various conditions.}
	\label{fig: henry}
\end{figure}


\subsection{Evaluation with Able-bodied Participants}
\label{ssec:eval_able}
\paragraph{\textbf{9 Able-bodied Participants}}
We recruited 9 able-bodied participants and performed evaluations in the Healthcare Robotics Lab between April 28th and May 12th, 2017. All participants were Georgia Tech students consisting of 1 female and 8 males, aged 22-29. We divided the participants into 3 groups of equal size where each group of participants used a different utensil and type of food: \textit{cottage cheese and silicone spoon}, \textit{watermelon chunks and metal fork}, and \textit{fruit mix and plastic spoon}. Each participant performed 60 non-anomalous feeding executions spread across three sessions. Each session lasted less than one hour and consisted of 20 non-anomalous feeding executions. The participants performed 540 non-anomalous executions and 19 extra executions. The participants also answered 11 post-experiment questions (five-point Likert type questionnaire items) after the experiment. Participants recorded their perceived comfort, risk, and convenience of the \textit{active feeding} system through the post-experiment questionnaire.


\paragraph{\textbf{Long-term Self Evaluation}}
We also designed a long-term evaluation to observe the system's daily assistance capability. The first author, an able-bodied participant, conducted a total of 428 feeding executions across 22 days between April 3rd and July 28th, 2017. The participant ate 6 types of foods (i.e., yogurt, rice, fruit mix, watermelon chunks, cereal, and cottage cheese) and used 5 utensils (i.e.,small/large plastic spoons, a silicone spoon, and plastic/metal forks) for lunch or dessert in the Healthcare Robotics Lab at Georgia Tech. The participant used the touch-based GUI throughout all feeding trials, with each feeding session lasting at most 30 minutes. Fig.~\ref{fig: self_study_food} shows the 6 examples of foods we used in this evaluation. For each session, the robot would continually provide feeding assistance until the robot had successfully fed all food from the bowl. In Section~\ref{sec:6}, we discuss the success and failure of the various feeding tasks. 

\begin{figure*}[t]
	\centering
 	  \includegraphics[width=180mm]{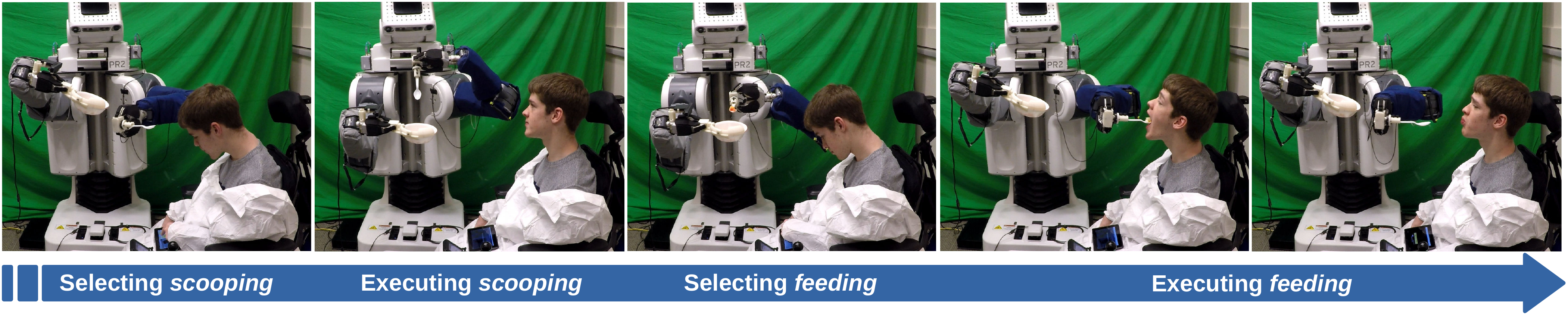}
	\caption{Evaluation with a person with motor impairments in an experiment room at Georgia Tech.}
	\label{fig: feeding_eval_with_disabilities}
\end{figure*} 

\begin{figure*}[t]
\centering
  \includegraphics[width=0.8\textwidth]{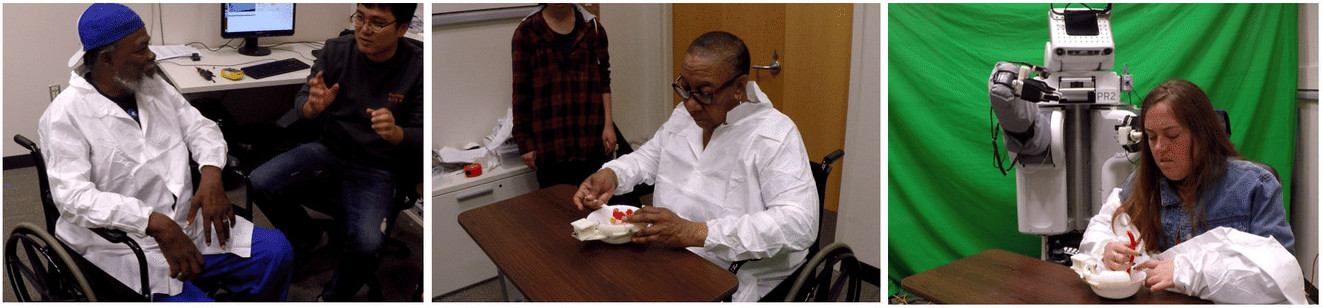}
\caption{In-lab evaluation with people with motor impairments, who have a self-feeding disability. The evaluation includes the pre/post interviews and self-feeding studies as well as the running tests.}
\label{fig:in_lab_evaluation}       
\end{figure*}


\subsection{Evaluation with People with Motor Impairments}

\paragraph{\textbf{In-home Evaluation}}
We also deployed the system to Henry's home in California, USA (see Fig.~\ref{fig: henry}). As a part of the monitoring system evaluation for \cite{park2017class}, he used the meal-assistance system for six sessions from February 11th through February 13th, 2017. He performed over 130 feeding executions under various light conditions (e.g., morning, evening, and night times). We designed this study to observe if he could use the system without assistance from experimenters or caregivers. For this evaluation, we used a distinct PR2 robot from Georgia Tech in his home and mounted the same equipment that we used in the laboratory. We used yogurt and a silicone spoon. During the evaluation, the robot held a bowl and a spoon and was located beside his wheelchair. He used a laptop with an off-the-shelf head tracker to move the mouse cursor and a one-button mouse to successfully control the robot through the web-based GUI. We did not provide Henry with a scripted action sequence and instead allowed him to freely use the system to eat yogurt using the Scooping, Wiping, and Feeding buttons. At the end of our evaluation, we asked Henry to fill out a survey with 22 questions (five-point Likert type questionnaire items) based on~\cite{weiss2009usus}, and 2 open-ended questions. Further details are described in \cite{park2017class}.

\paragraph{\textbf{8 People with Motor Impairments}}
After confirming the usability and safety of the system through the preliminary evaluations, we recruited 8 potential end users over the course of 5 months starting in November, 2017. All 8 participants had motor impairments which made it difficult for them to eat by themselves using hand-held utensils. 4 participants were male, and 4 participants were female. The age range was 23-72 ($\text{avg.}=49.4, \text{std.}=21.4$). Participants had self-reported motor impairments, but were comfortable with operating a touchscreen tablet. As shown in Fig.~\ref{fig: feeding_eval_with_disabilities}, they participated in the study on their power wheelchairs and fed themselves through the tablet able-bodied participants used. For each participant, we conducted 1 session lasting approximately 2 hours. After safety training and 3 practice trials, the participants were asked to use the robot to freely feed themselves 10 spoons of yogurt and 10 forks of mixed fruit using the web-based GUI. Experimenters refilled the bowl with food after every 5 feeding executions to ensure the robot consistently served an adequate amount of food. Our system gradually takes smaller amounts of food as the food in the bowl is reduced. We discuss insights about providing an adequate amount of food in Section~\ref{ssec:insight}. At the end of the experiment, we administered questionnaires based on the NASA TLX subjective workload measure \cite{hart1988development} and five-point Likert type questionnaire items as well as 2 open-ended questions. In addition to the formal questionnaires, we engaged participants in free discussion about the design of the meal-assistance system to gain insights from the targeted user group. Fig.~\ref{fig:in_lab_evaluation} shows photographs from this in-lab evaluation.


\begin{table*}[t] 
\caption{Five-point Likert type questionnaire items for 9 able-bodied participants. The last column provides the average and standard deviation of scores with 1=strongly disagree, 2=disagree, 3=neither, 4=agree, and 5=strongly agree. Results are considered to have reasonable accuracy when relative standard error (RSE) is 25\% or less.}
\begin{center}
\begin{tabular}{l c c c}
\toprule
\multirow{2}{*}{Questionnaires} & \multicolumn{3}{c}{Score} \\
\cmidrule(lr){2-4}
& Avg. & Std. & RSE \\
\midrule
    I am familiar with engineering. & 4.78 & 0.67 & 4.65\% \\
    I am familiar with robotic applications. & 4.89 & 0.33 & 2.27\% \\
    I successfully ate food using the system. & 4.67 & 1.00 & 7.14\% \\
    I am satisfied with using the system. & 3.89 & 0.93 & 7.95\% \\
    The system was easy-to-use. & 4.00 & 1.00 & 8.33\% \\    
    I felt safe while using the system. & 4.22 & 0.83 & 6.58\% \\        
    I was comfortable while using the system. & 4.33 & 0.71 & 5.44\% \\        
    The system delivered an adequate amount of food. & 3.00 & 1.58 & 17.57\% \\        
    The system delivered food with adequate speed. & 3.11 & 1.69 & 18.12\% \\        
    The system accurately placed food in my mouth & 4.00 & 0.87 & 7.22\% \\        
    The system provides sufficient safety tools or functions to prevent hazards. & 4.56 & 0.89 & 6.45\% \\        
\bottomrule
\end{tabular}
\label{table: survey_able_bodied}
\end{center}
\end{table*}

\section{Results} \label{sec:6}

\subsection{Evaluation with Able-bodied Participants} 
In this preliminary evaluation, we confirmed the usability and safety of this prototype meal-assistance system with 9 able-bodied participants. Table \ref{table: survey_able_bodied} shows 11 five-point Likert type questionnaire items. Based on scores from the questionnaire, we found that the 9 able-bodied participants believed that they were able to successfully feed themselves using the system with an average score of 4.67 (out of 5) and 7.14\% relative standard error (RSE)\footnote{An RSE less than 25\% indicates that the answer shows reasonable accuracy.}. Participants also reported that the system was safe and easy-to-use with scores of 4.22 and 4.0, respectively. Note that the participants were mostly familiar with robotic applications with a score of 4.89, so the answers may not be similar to the acceptance of the end-user group. An interesting result is that the participants neither agreed nor disagreed that the system provided an adequate amount of food or that the system delivers food with adequate speed. However, as discussed in the following section, people with motor impairments mostly agreed that the speed was adequate, despite the speed of the system being approximately the same. 


\begin{figure}[t]
	\centering
 	  \includegraphics[width=80mm]{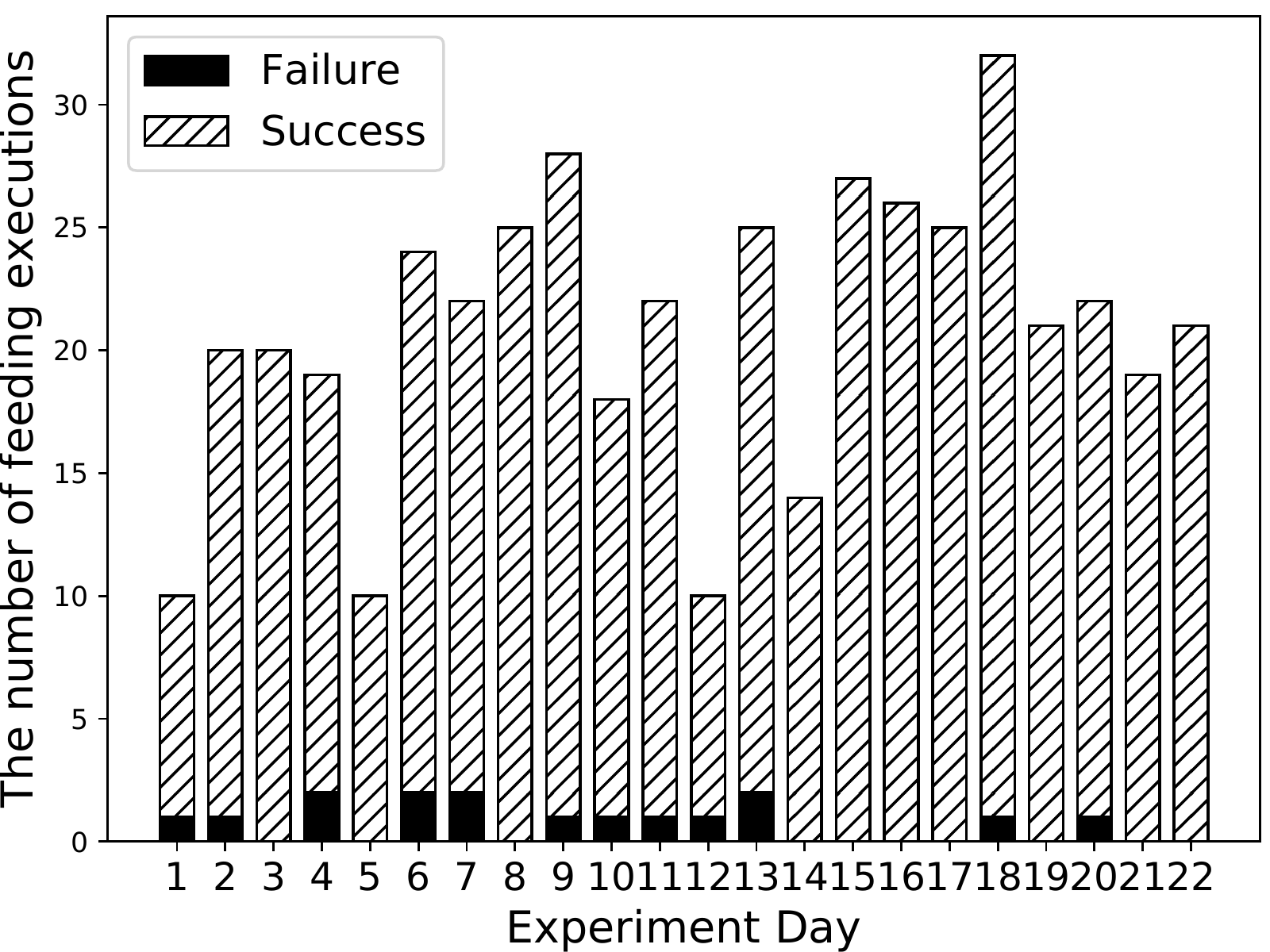}
	\caption{The number of daily successful and failed executions during the long-term self evaluation.}
	\label{fig: daily_failures}
\end{figure}

Fig.~\ref{fig: daily_failures} shows the number of daily successful and failed executions during the long-term self evaluation. Each day, the participant selected a utensil from the 5 utensils and a type of food to be fed: yogurt (6 days), rice (6 days), fruit (6 days), cereal (1 day), cheese (3 days). The participant also recorded the different kinds of failures that occurred during the experiment. Throughout this evaluation, the system resulted in an average of 16 ($3.6\%$) feeding failures out of a total of 444 executions due to \textit{a camera fault}, \textit{false alarms from the execution monitor}, \textit{tool collisions by system fault}, \textit{a system freeze}, and \textit{unknown reasons}. Through this long-term evaluation, we were able to stabilize the system and improve safety before proceeding to the evaluation with people with motor impairments.


\begin{table*}[t] 
\caption{Five-point Likert type questionnaire items for 8 people with motor impairments. The last column provides the average and standard deviation of scores with 1=strongly disagree, 2=disagree, 3=neither, 4=agree, and 5=strongly agree.}
\begin{center}
\begin{tabular}{l c c c}
\toprule
\multirow{2}{*}{Questionnaires} & \multicolumn{3}{c}{Score} \\
\cmidrule(lr){2-4}
& Avg. & Std. & RSE \\
\midrule
    I feel comfortable using my current feeding system. & 2.88 & 1.64 & 20.19\% \\
    I feel independent using my current feeding system. & 3.25 & 1.49 & 16.19\% \\
    I expect this meal-assistance system to increase the independence of the user. & 4.14 & 0.90 & 7.68\% \\
    I expect this meal-assistance system to be satisfactory. & 4.38 & 0.74 & 6.01\% \\
    I expect this meal-assistance system to be comfortable. & 4.38 & 0.74 & 6.01\% \\    
    I am comfortable with using technology. & 4.57 & 0.53 & 4.13\% \\     
    I felt comfortable using the meal-assistance system. & 4.75 & 0.46 & 3.45\% \\        
    I felt independent using the meal-assistance system. & 4.50 & 1.07 & 8.40\% \\        
    The meal-assistance system provided significant help in eating. & 4.88 & 0.35 & 2.56\% \\        
    The meal-assistance system successfully accomplished tasks. & 4.38 & 0.52 & 4.18\% \\        
    The meal-assistance system was simple and easy to use. & 4.75 & 0.71 & 5.26\% \\    
    I felt safe while using the meal-assistance system. & 4.50 & 0.76 & 5.94\% \\    
\bottomrule
\end{tabular}
\label{table: survey_disabilities}
\end{center}
\end{table*}

\begin{table*}[t] 
\caption{Questionnaire items based on NASA-TLX for 8 people with motor impairments. Participants chose a number between 1 and 20 for each question. 1 indicates ``Very Low" for Mental, Physical, Temporal, Effort, and Frustration, and ``Perfect" for Performance. 20 indicates ``Very High" for Mental, Physical, Temporal, Effort, and Frustration, and ``Failure" for Performance.}
\begin{center}
\begin{tabular}{l l c c c}
\toprule
&\multirow{2}{*}{Questionnaires} & \multicolumn{3}{c}{Score} \\
\cmidrule(lr){3-5}
&& Avg. & Std. & RSE \\
\midrule
    Mental demand& How mentally demanding was the task? & 2.88 & 2.03 & 23.55\% \\
    Physical demand& How physically demanding was the task? & 4.00 & 5.01 & 41.79\% \\
    Temporal demand& How hurried or rushed was the pace of the task? & 2.25 & 2.38 & 35.17\% \\
    Performance& How successful were you in accomplishing what you were asked to do? & 4.50 & 4.34 & 32.17\% \\
    Effort& How hard did you have to work to accomplish your level of performance? & 3.38 & 4.50 & 44.46\% \\    
    Frustration& How insecure, discouraged, irritated, stressed, and annoyed were you? & 5.25 & 5.31 & 33.73\% \\     
\bottomrule
\end{tabular}
\label{table: survey_nasatlx}
\end{center}
\end{table*}


\subsection{Evaluation with People with Motor Impairments}
For the in-home evaluation, an end user, Henry Evans, successfully fed himself with the robot for all consecutive trials. 
Henry reported that he found the system to be effective, safe, and easy to use. Further details are described in \cite{park2017class}. We discuss new observations in Sec.~\ref{ssec_analysis} and \ref{sec:7}.

We conducted evaluations with 8 people with motor impairments. The system succeeded at feeding 99 times out of a total of 100 attempts (task success). The only failure was due to a participant's accidental stop button click. The participants also answered that the system successfully fed them with an average score of 4.52 out of 5 (user success). Participants' responses suggest that the robotic system could increase their perceptions of comfort and independence over their current feeding systems. They gave average scores of 2.88 and 3.25 to comfort and independence questions about their current feeding systems versus 4.75 and 4.5 for the robotic system (see Table \ref{table: survey_disabilities}). The participants agreed that the system helps feeding significantly with an average score of 4.88. Further analyses are in Sec.~\ref{ssec_analysis}. We additionally measured NASA TLX subjective workload to assess the participants' mental, physical, temporal, effort, and frustration while using the system (see Table~\ref{table: survey_nasatlx}). The result indicates that our system requires an overall low workload. In particular, the simple interface of system resulted very low mental demand. However, the average of frustration is higher than other demands. We leave the investigation of the relation between the frustration and the system as our future work.  
%



\begin{figure}[t]
	\centering
 	  \includegraphics[width=70mm]{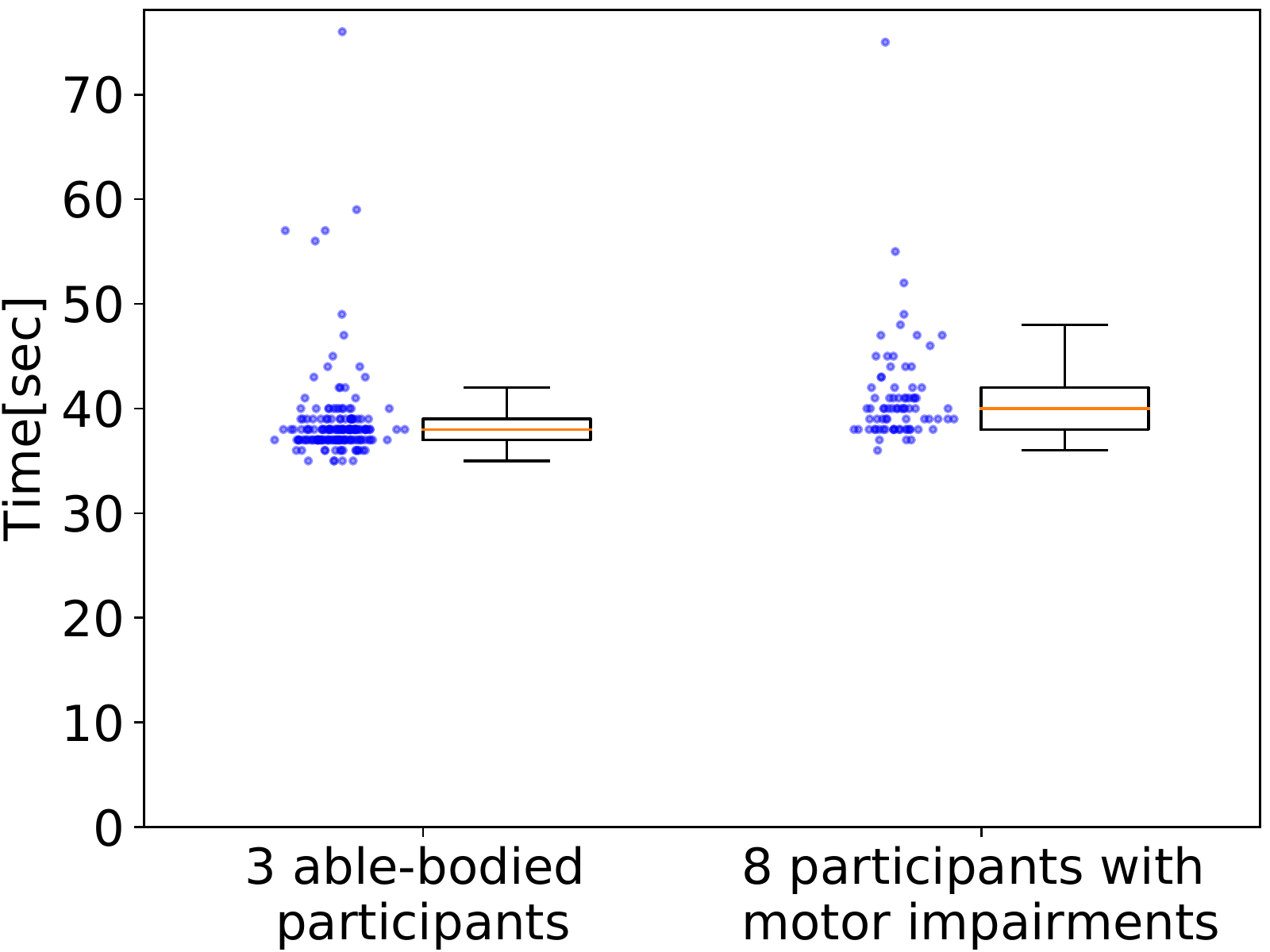}
	\caption{Distributions of meal-assistance completion time from 3 able-bodied participants and 8 people with motor impairments ($p\text{-value}=0.0004$ with a two-sided Welch's $t$-test). The participants ate yogurt with a silicone spoon.}
	\label{fig: feeding_duration}
\end{figure}


\subsection{Analyses} \label{ssec_analysis}
We compared evaluation results from able-bodied people and people with motor impairments in terms of five selected factors.

\paragraph{\textbf{Completion Time}}
Fig.~\ref{fig: feeding_duration} shows the distributions of meal-assistance completion time from 3 able-bodied participants and 8 participants with motor impairments, where the 3 able-bodied participants are the group who used the same type of food and utensil (i.e., yogurt and a silicone spoon) as participants with motor impairments did. In this graph, each participant ate yogurt using a silicone spoon. The completion time is an elapsed time that has passed between the click of the scooping button and the end of feeding motion including spoon wiping executed by individuals' preference. There was no speed-relevant system change between two experiments. Able-bodied participants and those with motor impairments took about $39\pm 4.5\sec$ and $41 \pm 5.2\sec$, respectively. We performed Welch's $t$-test, also known as unequal variances $t$-test, to test whether two samples differ significantly. The test resulted in $p\textit{-value} = 0.0004$, which indicates the difference between two participant groups' completion times is statistically significant. A likely cause is the difficulty of GUI manipulation due to upper-limb impairments.

Henry Evans also took about 78 seconds for one time of meal assistance that is about 39 seconds longer than the able-bodied participants' duration. A likely cause was from mouse pointing time using the head tracker and the use of the wiping subtask, since Henry Evans used it 0.85 times per meal assistance but other participants mostly did not use it. Note that the wiping subtask usually took 17 seconds and there were adjustments in motion between the in-home and in-lab evaluations. However, there was no change in speed.

\begin{figure}[htbp]
	\centering
	\begin{subfigure}[t]{0.47\textwidth}
		\centering
		\includegraphics[trim={0cm, 0cm, 0.5cm, 0cm}, clip,width=85mm]{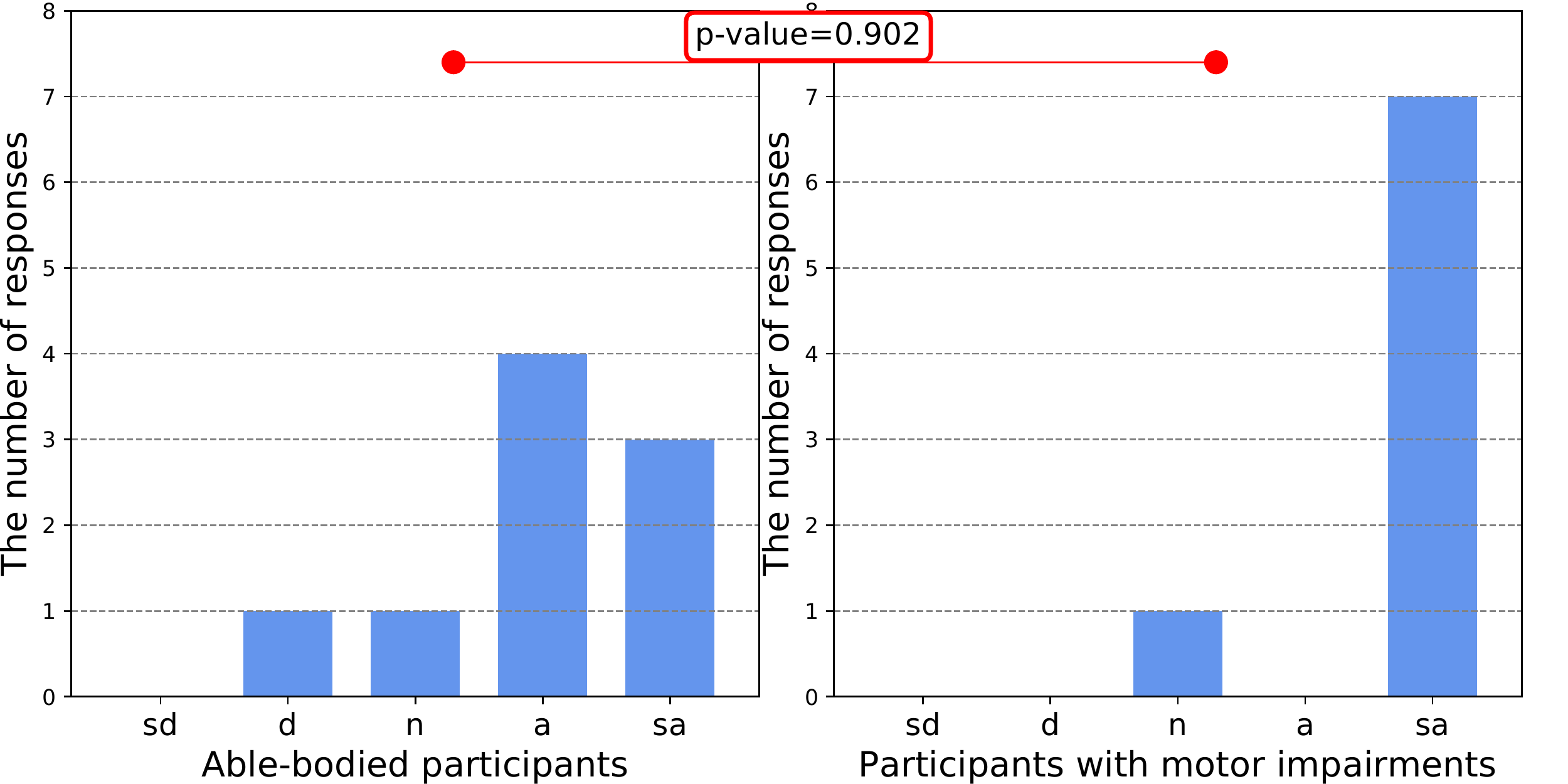}
        \caption{`The system is easy-to-use to performing self-feeding task.'}
		\label{fig: easy_to_use}
    \end{subfigure}
	\begin{subfigure}[t]{0.47\textwidth}
		\centering
 	  	\includegraphics[trim={0cm, 0cm, 0.5cm, 0cm}, clip,width=85mm]{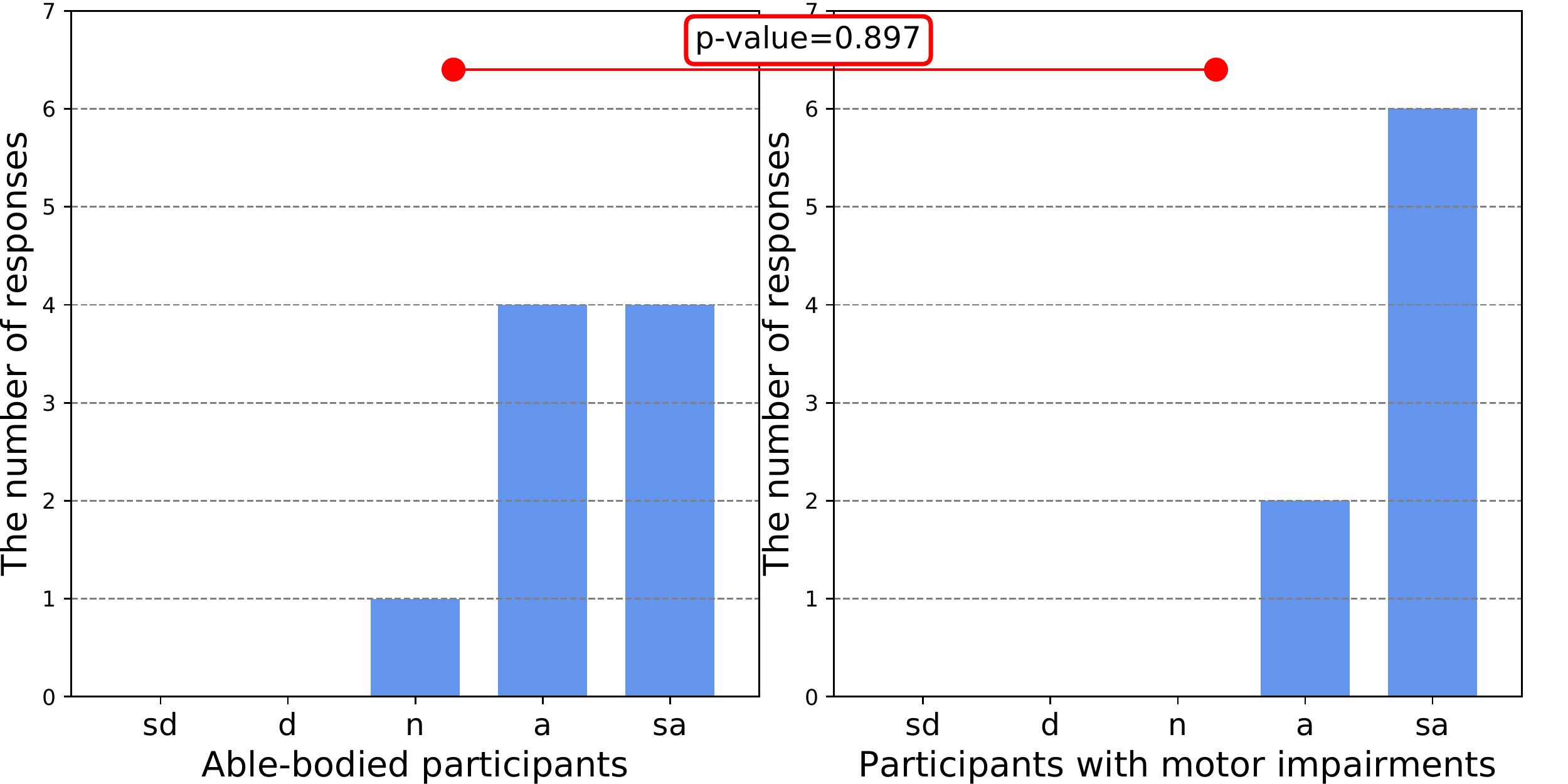}
        \caption{`The system is comfortable to performing self-feeding task.'}
		\label{fig: comfort}
    \end{subfigure}
	\begin{subfigure}[t]{0.47\textwidth}
		\centering
 	  	\includegraphics[trim={0cm, 0cm, 0.5cm, 0cm}, clip,width=85mm]{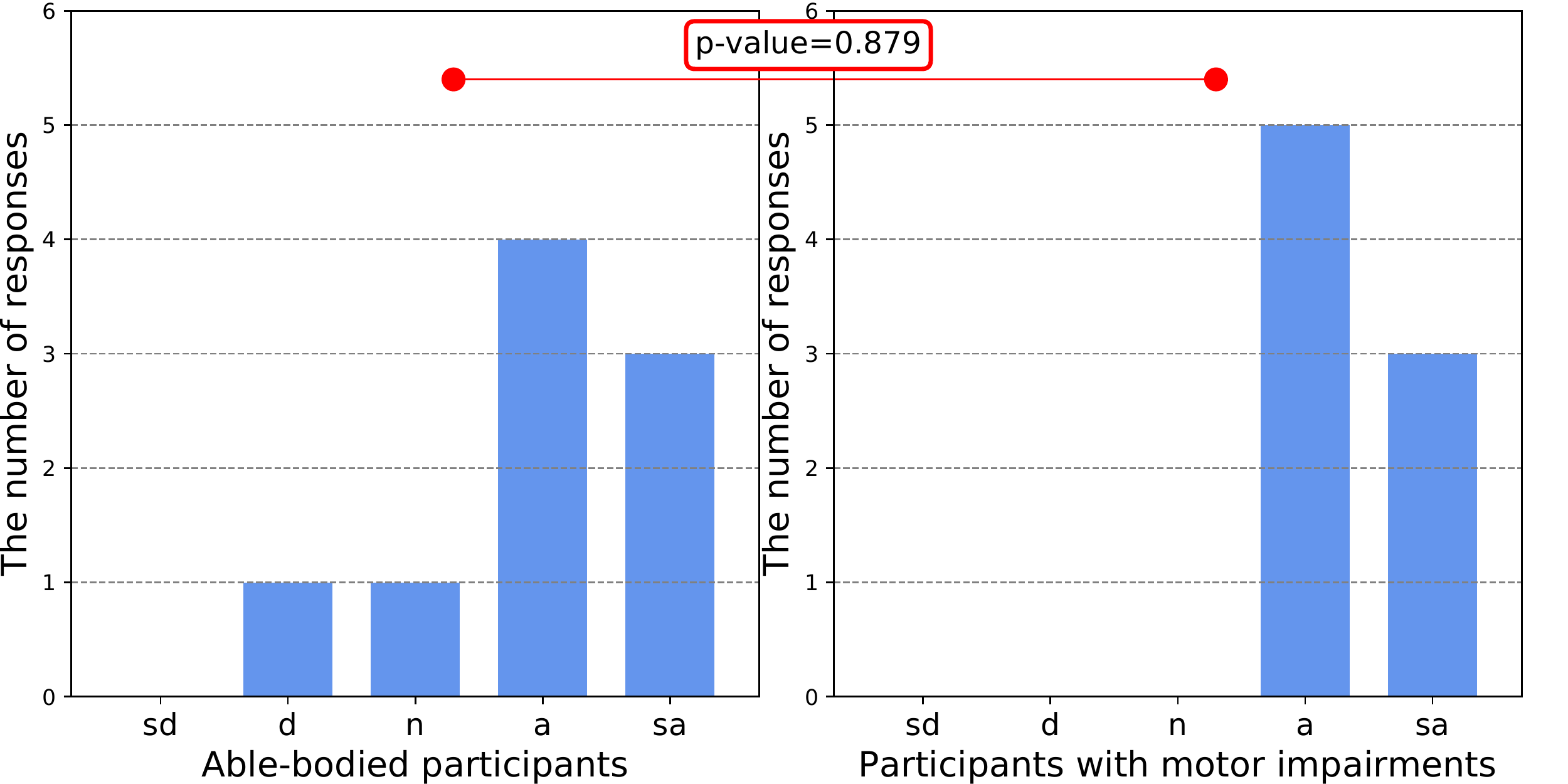}
        \caption{`The system is successful to performing self-feeding task.'}
		\label{fig: success}
    \end{subfigure}
	\begin{subfigure}[t]{0.47\textwidth}
		\centering
 		\includegraphics[trim={0cm, 0cm, 0.5cm, 0cm}, clip,width=85mm]{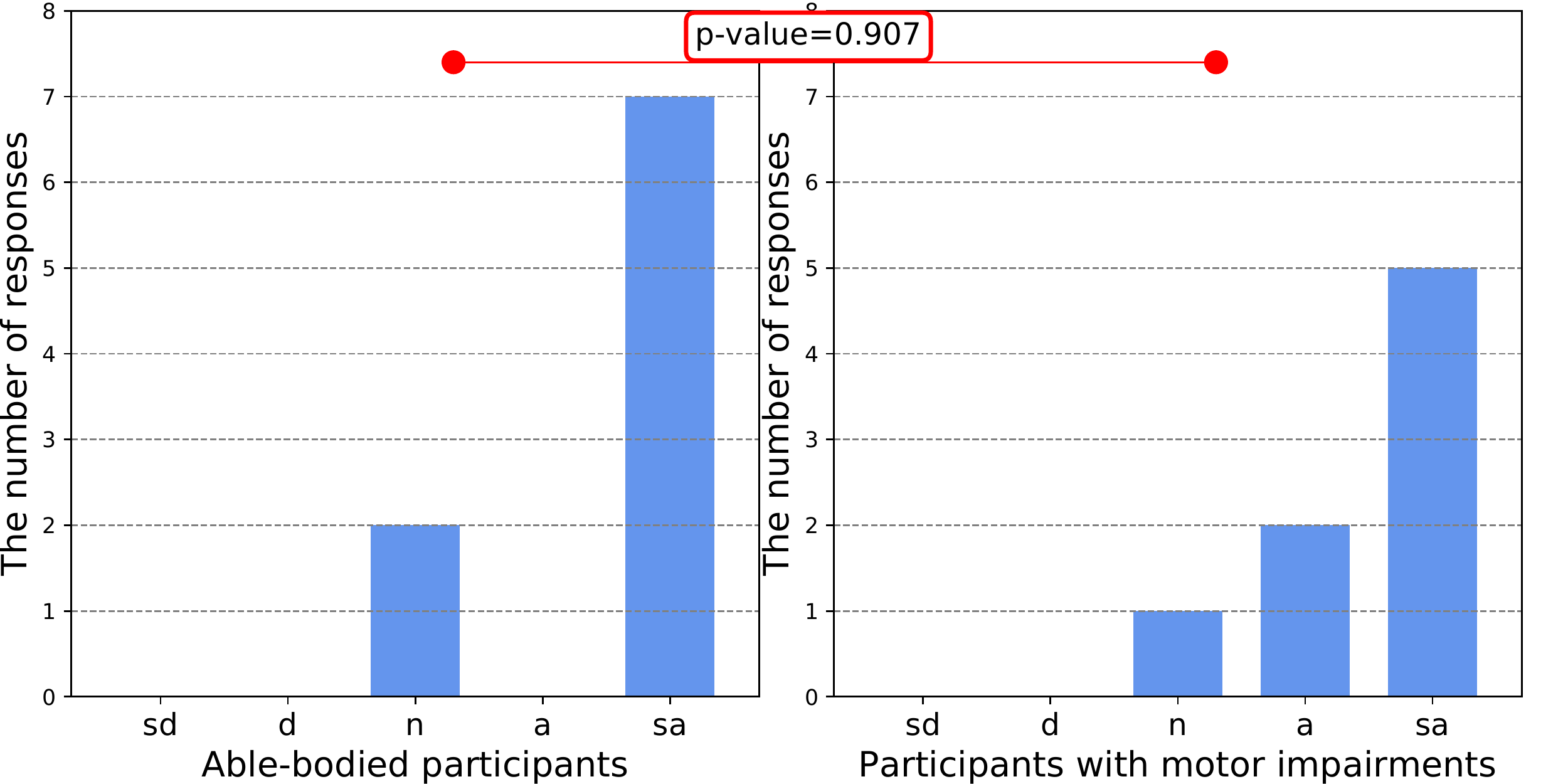}
        \caption{`The system is safe to performing self-feeding task.'}
		\label{fig: safety}
    \end{subfigure}
	\caption{9 able-bodied participants (Left) and 8 participants with motor impairments (Right) reported the level of agreement with the statement in each sub caption where sd=strongly disagree, d=disagree, n=neither, a=agree, and sa=strongly agree. $p\text{-value}$ is computed from the two-sided Welch's $t$-test between two graphs.}
\end{figure}

\paragraph{\textbf{Ease of Use}}
To assess ease of use, we asked a question about ease of use to both able-bodied people and those with motor impairments. Fig.~\ref{fig: easy_to_use} shows their responses where the median responses are `agree' for able-bodied people and `strongly agree' for those with motor impairments. The Welch’s $t$-test resulted in $p\textit{-value}=0.901$, so we cannot conclude that their agreements significantly differ.

\paragraph{\textbf{Comfortable Assistance}}
We assessed agreement with `the system is comfortable to perform self-feeding task.' Fig.~\ref{fig: comfort} shows the participants' responses where the median responses are `agree' for able-bodied people and `strongly agree' for those with motor impairments. The Welch’s $t$-test resulted in $p\textit{-value}=0.897$, so we cannot conclude that their agreements significantly differ.

\paragraph{\textbf{Successful Assistance}}
Self-assessment of success was measure with the item `the system is successful to perform self-feeding task.' Fig.~\ref{fig: success} shows the participants' responses where the median responses are `agree' for both. The Welch’s $t$-test resulted in $p\textit{-value}=0.879$, so we cannot conclude that their agreements significantly differ.
\paragraph{\textbf{Safety}}
Safety concerns were measured with the statement `the system is safe to perform self-feeding task.' Fig.~\ref{fig: safety} shows the participants' responses where the median responses are `strongly agree' for both. The Welch’s $t$-test resulted in $p\textit{-value}=0.907$, so we cannot conclude that their agreements significantly differ.

Throughout the evaluations, our meal-assistance system successfully fed foods to both able-bodied participants and those with motor impairments. Participants agreed that the system comfortably, successfully, and safely provides the meal assistance with easy-to-use interface. Overall, our results suggest that it is feasible for general-purpose mobile manipulators to provide meal assistance.

\section{Discussion} \label{sec:7}

\subsection{Design Insights} \label{ssec:insight}
We discuss learned lessons and insights about the design of potential meal-assistance systems for people with motor impairments that result in a self-feeding disability.

\begin{itemize}
\item \textbf{Robot Appearance:} Participants responded that they felt safe after the evaluation. However, several participants were overwhelmed or intimidated at first by the large size of the robot (a PR2). A participant with motor impairments stated that she felt threatened when the thick arm approached her in the beginning of the evaluation. Another participant with motor impairments said that a robot which is bigger than herself is intimidating. As for possible solutions to this problem, one participant with motor impairments suggested to make the robot look more ``friendly", such as by using more natural colors rather than the current metallic exterior.
\item \textbf{User Interface (UI):} In our study with 8 people with disabilities, the users were required to use the same touchscreen device. Having alternative devices or methods of providing input to the device would be useful. A participant with numbness in her fingers often had trouble using the touchscreen, and preferred physical buttons. A number of participants clicked the touchscreen multiple times due to no button feedback. However, another participant did not have much strength in her hands, and therefore preferred the touchscreen to other methods. Our system's web-based interface can support alternatives, as demonstrated by Henry Evans. 
\item \textbf{Slicing food:} Many participants with motor impairments stated that slicing/cutting food is something that could be improved about their current feeding method, since they have difficulty applying sufficient force and dexterously manipulating knives. An assistive robot that slices food has the potential to facilitate independent eating with a variety of foods and reduce caregiver burden.  
\item \textbf{Amount of Food:} Serving an adequate amount of food would help to increase the satisfaction of users and the efficiency of task executions. In the evaluations, the amount of food the robot scooped/stabbed varied. An excessive amount of food was usually connected to food spills while delivering food, and would sometimes require more than one task execution to finish all of the food on the spoon. At other times, the system scooped a very small amount of solid food (e.g., fruit) or failed to scoop. 
\item \textbf{Speed:} Providing adjustable speed may be beneficial. The desired speed varied across participants and participant groups. Able-bodied participants neither agreed nor disagreed that the current system's speed (about \SI{40}{sec} per cycle) was adequate. In contrast, the responses from people with motor impairments indicated they were satisfied with the speed on average, although Henry Evans desired a faster rate of feeding. Henry's preference may be due to him being an expert user.

\item \textbf{Delivery Motion:} 
Tilting a spoon when it is leaving a user's mouth can be beneficial. Depending on the spoon's depth and the user's disability, a horizontal retracting motion may not be sufficient to move all of the food from the spoon into the user's mouth. For example, in the first author's long term evaluation, the participant had difficulty eating all food scooped by the large plastic spoon in Fig.~\ref{fig: feeding_tool}. Adding a human-like spoon feeding motion (i.e., tilting and retracting) enabled the participant to comfortably eat all food from the spoon with ease.

\item \textbf{Emergency Alarm:} We invited participants who do not have difficulties eating the selected food (e.g., dysphagia). On the other hand, the participant, Henry Evans, for in-home evaluation has severe motor impairments, is unable to speak, and has difficulty chewing and swallowing food. To reduce any risk for aspiration, we made him sit upright and provided yogurt as a pureed food following known aspiration precautions. For users who cannot resolve aspiration issues by themselves, assistive robots potentially need to provide an emergency alarm function to alert nearby caregivers in the future.
\end{itemize}

\subsection{Assistive Robots}
\begin{itemize}
\item The cost of a general-purpose robot is decreasing rapidly due to growing commercial interest. In 2010, Willow Garage released a wheeled dual manipulator, the PR2, for \$400,000 \cite{pr2}. In 2017, PAL Robotics started to offer the noticeably lower cost 14-DoF mobile manipulator, TIAGo Steel, for \euro{49,900} \cite{tiago}, while the Obi feeding robot, a single-purpose robot, still costs \$5,950 \cite{obi}. In addition, open-source software and hardware have rapidly improved in availability and quality over the last decade. The development of assistive applications using general-purpose mobile manipulators could potentially be economically affordable in the near future. It could help address challenges associated with the aging population, rising healthcare costs, and shortages of healthcare workers in the United States and other countries. These inventions also have the potential to reduce family caregivers' prolonged stress, physical demands, and decline in quality of life while reducing the healthcare costs for households with people with disabilities.
\item The system we have presented is focused on assistance rather than rehabilitation. Automated assistance can enable people with diverse disabilities to perform daily activities by themselves. However, robotic assistance could also potentially discourage people from using their physical abilities in activities that they could perform with less assistance. There is also the potential for robotic assistance to affect peoples' rehabilitation status or worsen their health through reduced physical activity. These are risks that merit attention as robotic assistance becomes more common. In addition to assistance, general-purpose robots could potentially help during rehabilitation or modify their assistive actions to encourage exercise by users. 
\end{itemize}


\section{Conclusion} \label{sec:8}
We introduced a proof-of-concept of meal-assistance system using a general-purpose mobile manipulator, a PR2 robot. The system can perform three independent subtasks: scooping/stabbing, spoon wiping, and delivery, where a user can command a preferred subtask via a web-based GUI. Unlike conventional feeding devices, our novel system design enabled the mobile manipulator to provide visually-guided \textit{active feeding} assistance that autonomously delivers food inside a user's mouth after visually-guided scooping/stabbing of food. We also designed the system to provide safer assistance using various hardware and software tools including a high-level execution monitor. Overall, the design improved the accessibility and usability of the meal-assistance system for people with motor impairments that led to self-feeding disability.

We evaluated the system with total 10 able-bodied participants and 9 participants with motor impairments. In our evaluation with 9 able-bodied participants, the system successfully performed roughly 2,000 feeding executions with 5 utensils and 6 types of food items. Throughout our longer term self evaluation, we confirmed the safety and usability of the system. In our evaluation with the end-user group, the participants with motor impairments were able to use the system successfully to feed themselves. Their responses were generally positive and similar to those of able-bodied participants to questions about the ease-of-use, comfort, safety, and success of the system. We also deployed the system at Henry Evans' house in California, USA. He was able to use the system to feed himself successfully with 70 non-anomalous feeding executions for three days in his residential home setting. We demonstrated the feasibility of the new meal-assistance system. Finally, we shared design insights and lessons we learned through the design and evaluation. Our robot-assisted feeding system has the potential to reduce self-feeding limitations in people with motor impairments by providing them the support they need to perform feeding tasks.

\section{Appendix: Supplementary data} \label{sec:9}
The evaluation process and interview scenes are attached as a supplementary video.

\section*{Acknowledgements}
We thank Mark Killpack who provided the control library. This work was supported by NSF Award IIS-1150157, NIDILRR grant 90RE5016-01-00 via RERC TechSAge, and a Google Faculty Research Award. 

Dr. Kemp is a cofounder, a board member, an equity holder, and the CTO of Hello Robot, Inc., which is developing products related to this research. This research could affect his personal financial status. The terms of this arrangement have been reviewed and approved by Georgia Tech in accordance with its conflict of interest policies.

\bibliographystyle{model1-num-names}
\bibliography{feeding_system}   

\end{document}